\documentclass[10pt,journal,compsoc]{IEEEtran}
\ifCLASSOPTIONcompsoc
  \usepackage[nocompress]{cite}
\else
  \usepackage{cite}
\fi
\ifCLASSINFOpdf
  \usepackage[pdftex]{graphicx}
\else
  \usepackage[dvips]{graphicx}
\fi
\usepackage{amsmath}
\usepackage{amsthm}
\usepackage{algorithmic}
\usepackage{array}
\ifCLASSOPTIONcompsoc
 \usepackage[caption=false,font=footnotesize,labelfont=sf,textfont=sf]{subfig}
\else
 \usepackage[caption=false,font=footnotesize]{subfig}
\fi
\usepackage{fixltx2e}
\usepackage{dblfloatfix}
\ifCLASSOPTIONcaptionsoff
 \usepackage[nomarkers]{endfloat}
\let\MYoriglatexcaption\caption
\renewcommand{\caption}[2][\relax]{\MYoriglatexcaption[#2]{#2}}
\fi

\usepackage{url}

\usepackage[utf8]{inputenc} % allow utf-8 input
\usepackage[T1]{fontenc}    % use 8-bit T1 fonts
\usepackage{hyperref}       % hyperlinks
\usepackage{booktabs}       % professional-quality tables
\usepackage{amsfonts}       % blackboard math symbols
\usepackage{nicefrac}       % compact symbols for 1/2, etc.
\usepackage{microtype}      % microtypography
\usepackage{xcolor}         % colors
\usepackage{makecell}
\usepackage{threeparttable}
\usepackage{gensymb}
\usepackage{bm}
\usepackage{textcomp, gensymb}
\usepackage{graphics}
\usepackage{multirow}
\usepackage{multicol}
\usepackage{enumitem}
\usepackage[ruled,vlined]{algorithm2e}
\usepackage{float}
\usepackage{subfig}
\usepackage{tabularx}
\usepackage{longtable}
\usepackage{tabu}
\usepackage{dsfont}
\usepackage{amsmath,amssymb}
\usepackage{pifont}% http://ctan.org/pkg/pifont
\usepackage{listings,lstautogobble}
\newcommand{\etal}{\textit{et al}.~}
\newcommand{\ie}{\textit{i}.\textit{e}.}
\newcommand{\eg}{\textit{e}.\textit{g}.}

\newcommand{\red}[1]{{\color{black} #1}}
\newcommand{\bl}[1]{{\color{black} #1}}

\newtheorem{definition}{Definition}

\DeclareMathOperator*{\argmin}{arg\,min}

\usepackage{etoolbox}
\begin{document}

\title{DeepDC: Deep Distance Correlation \\as a Perceptual Image Quality Evaluator}

\author{Hanwei~Zhu,
        Baoliang~Chen,
        Lingyu~Zhu, 
        Shiqi~Wang,~\IEEEmembership{Senior~Member,~IEEE},
        ~and Weisi~Lin,~\IEEEmembership{Fellow,~IEEE}% <-this % stops a space
\IEEEcompsocitemizethanks{
\IEEEcompsocthanksitem H. Zhu, B. Chen, L. Zhu, and S. Wang are with the Department
of Computer Science, City University of Hong Kong, Kowloon, Hong Kong. (emails: \{hanwei.zhu blchen-6, lingyzhu-c\}@my.cityu.edu.hk, shiqwang@cityu.edu.hk).
\IEEEcompsocthanksitem W. Lin is with the School of Computer Science and Engineering, Nanyang Technological University, Singapore. (email: wslin@ntu.edu.sg).
}
\thanks{Corresponding author: Shiqi Wang.}
}

\markboth{}{}


\IEEEtitleabstractindextext{%
\begin{abstract}

ImageNet pre-trained deep neural networks (DNNs) are successful for high-level computer vision tasks and also show notable transferability for building effective image quality assessment~(IQA) models. Such a remarkable byproduct has often been identified as an emergent property in previous studies. In this work, we attribute such capability to the intrinsic \textit{texture-sensitive} characteristic of pretrained DNNs that classify images based on texture features. We fully exploit this characteristic to develop a novel full-reference IQA~(FR-IQA) model, which is exclusively based on the features of the pre-trained DNNs. Specifically, we compute the distance correlation, a highly promising yet relatively under-investigated statistic, between reference and \bl{distorted} images in the deep feature domain to develop the FR-IQA model. The distance correlation is computed as the \red{ratio} of their distance covariance to the product of their distance standard deviations. For the discrete observations (\ie, deep features), the distance covariance has a closed-form solution that computes the inner product of the double-centered distance matrixes. This enables the quantification of both linear and nonlinear feature relationships, which is far beyond the widely used first-order and second-order statistics~(\ie, mean, variance, and Gram matrix) in the feature space. We conduct comprehensive experiments to demonstrate the superiority of the proposed quality model on five standard IQA datasets, one perceptual similarity dataset, two texture similarity datasets, and one geometric transformation dataset.
Moreover, we optimize the proposed model to generate a broad spectrum of texture patterns, \bl{by treating the model as} the style loss function for neural style transfer (NST). Extensive experiments demonstrate that the proposed texture synthesis and NST \bl{methods} achieve the best quantitative and qualitative results. We release our code at \url{https://github.com/h4nwei/DeepDC}.

\end{abstract}

}

\maketitle

\IEEEdisplaynontitleabstractindextext
 
\IEEEpeerreviewmaketitle

\IEEEraisesectionheading{\section{Introduction}\label{sec:introduction}}
The primary goal of objective image quality assessment (IQA) is to automatically predict the perceptual visual quality, providing a cost-effective alternative for the cumbersome subjective user study~\cite{lin2011perceptual,zhai2020perceptual,duanmu2021quantifying}. The advancement of IQA models holds a pivotal position in the realms of human visual perception and computational vision systems. These models are broadly utilized to gauge the performance of image processing systems and to augment the optimization of a diverse range of practical applications, including but not limited to image compression, restoration, and rendering~\cite{yang2015perceptual,gu2015analysis,li2017quality,zhang2021plug}. The mean squared error~(MSE) and the structural similarity index~(SSIM) have evolved to become the acknowledged yardsticks in the field of image processing owing to their inherent simplicity, clear physical interpretation, and mathematical convenience for optimization~\cite{wang2004image}. 

Recently, deep neural networks~(DNNs) trained on ImageNet have exhibited significant perceptual quality assessment capability, which has been extensively utilized in the development of perceptual quality  models~\cite{zhang2018unreasonable,ding2020image,ADISTS,heusel2017gans,kligvasser2021deep,liao2022deepwsd,Manoj2022Do}. Zhang~\etal interpreted the perceptual quality assessment capability derived from image classification task as an emergent property~\cite{zhang2018unreasonable}. 
Subsequent studies have indicated that the high classification accuracy of  DNNs trained on ImageNet largely originates from their intrinsic \textit{texture-sensitive} characteristic, whereby images are classified based on superficial texture cues~\cite{geirhos2018imagenet,hermann2020origins}.
Considering that texture statistics, such as mean, variance, and Gram matrix, are essential components of full-reference IQA~(FR-IQA) model~\cite{heusel2017gans,ding2020image,kligvasser2021deep}, this \textit{texture-sensitive} property may provide a potential explanation for the effectiveness of the pre-trained~\footnote{Throughout this paper, we refer to the model trained on ImageNet~\cite{deng2009imagenet}.}  DNNs in evaluating perceptual quality. However, the full exploitation of this characteristic to build a more robust FR-IQA model still warrants further investigation.

\begin{figure*}
    \centering
    \captionsetup{justification=centering}
    \subfloat[DISTS $\downarrow$ / DeepDC $\downarrow$]{\includegraphics[scale=0.5,width=0.195\textwidth]{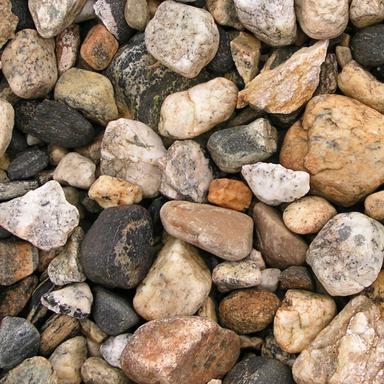}}\hskip.2em
    \subfloat[0.215 / 0.054]{\includegraphics[scale=0.5,width=0.195\textwidth]{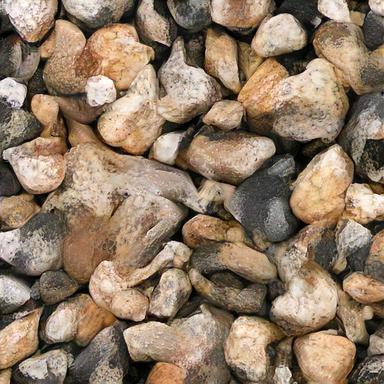}}\hskip.2em        
    \subfloat[0.239 / 0.071]{\includegraphics[scale=0.5,width=0.195\textwidth]{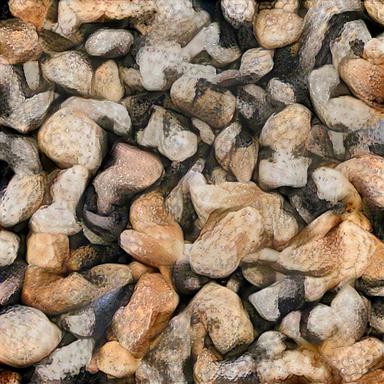}}\hskip.2em    
    \subfloat[0.213 / 0.037]{\includegraphics[scale=0.5,width=0.195\textwidth]{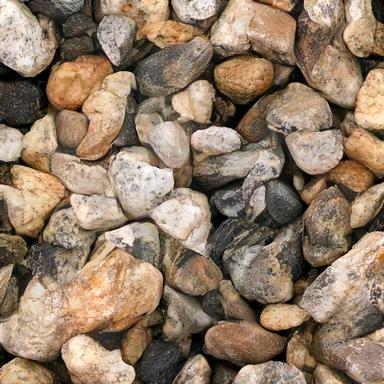}}\hskip.2em    
    \subfloat[\textbf{0.207} / \textbf{0.028}]{\includegraphics[scale=0.5,width=0.195\textwidth]{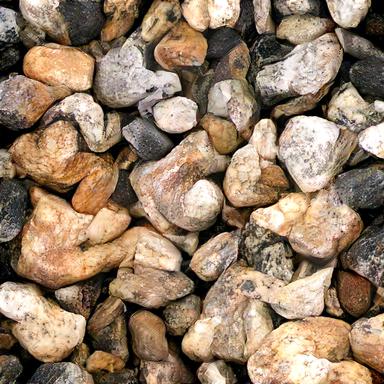}}
    \caption{Illustration of the comparisons of DeepDC against texture synthesis methods. (a) Reference image; (b) Gram matrix~\cite{gatys2015texture}; (c) Mean value~\cite{ding2020image}; (d) Sliced Wasserstein distance~\cite{heitz2021sliced}; (e) DeepDC \textit{Ours}. \red{A lower score of DISTS~\cite{ding2020image} and DeepDC indicates superior visual quality.}}\label{fig:sample_texture}
\end{figure*}

\begin{figure*}
    \centering
    \captionsetup{justification=centering}
    \subfloat{\includegraphics[trim={0 1cm 0 0}, scale=0.5,width=0.98\textwidth]{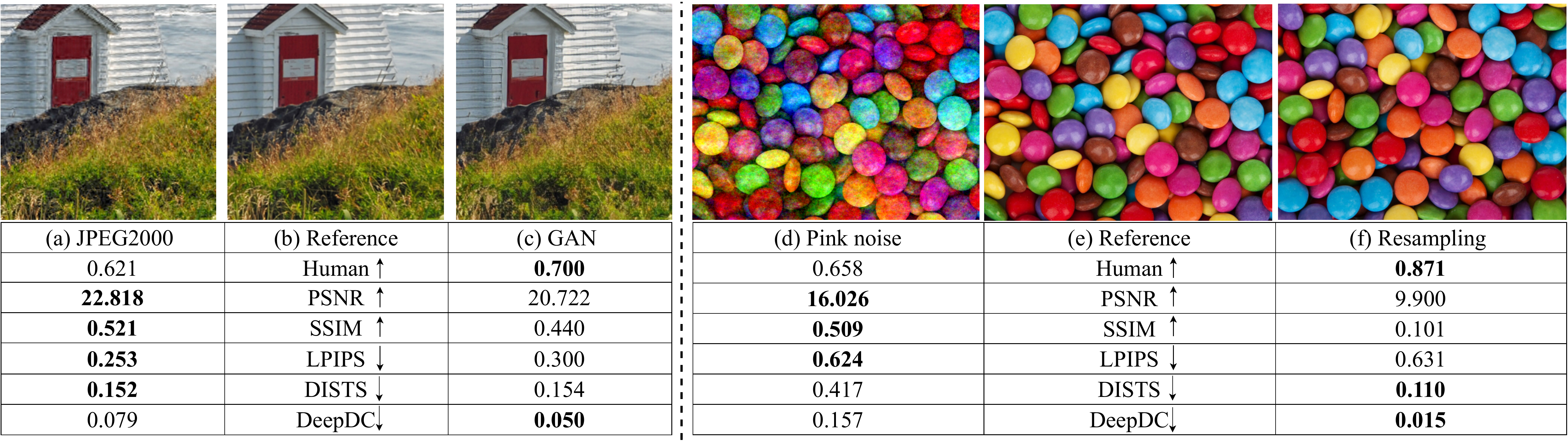}}\\
    \caption{Illustration of the contrastive preference by humans and the FR-IQA models.  Left: Human and DeepDC prefer the image (c) from GAN-baed method~\cite{navarrete2018multi} over the JPEG2000 distorted image (a), but the existing FR-IQA method (\ie, PSNR, SSIM, LPIPS, and DISTS) prefer image (a) over (c); Right: Human, DISTS, and DeepDC prefer the resampling texture image (f) over the pink noise image (d), but the PSNR, SSIM, and LPIPS prefer image (d) over (f). The better quality scores are highlighted in boldface.}\label{fig:sample}
\end{figure*}

The fundamental principle of perceptual FR-IQA models based on pre-trained DNN features involves the aggregation of deep representations, followed by a feature comparison. The learned perceptual image patch similarity (LPIPS)~\cite{zhang2018unreasonable} calculates the spatial average of the corresponding residual deep representations, which are further regressed to the final quality scores. To capture more comprehensive statistics, the second-order moment~(Gram matrix) is used to build the deep self-dissimilarity~(DSD) model~\cite{kligvasser2021deep}. Ding~\etal computes both the mean and variance of the pre-trained deep features to unify deep image structure and texture similarity~(DISTS)~\cite{ding2020image, ADISTS}. The Fréchet inception distance~(FID) quantifies the similarity by calculating the Fréchet distance between multivariate Gaussians, defined by their means and covariance matrices of the Inception network~\cite{heusel2017gans}. Furthermore, comparisons of internal feature distributions, such as the deep Wasserstein distance~(DeepWSD)~\cite{liao2022deepwsd}, play another role in the prediction of perceptual visual quality. However, as validated by an analysis-by-synthesis example~\cite{julesz1962visual,portilla2000parametric} in Fig.~\ref{fig:sample_texture}, we find the widely used feature statistics and distributions are insufficient to fully capture texture information~\red{(\eg, color, size, shape, and orientation)}~in pre-trained DNNs. First-order and second-order moments, which primarily measure linear feature relationships, are constrained by their limited representation capability. As a result, the resulting FR-IQA models require extensive fine-tuning with mean opinion scores (MOSs) and exhibit limitations in handling imperceptible perturbations in comparison to human perception in psychophysics. One example can be observed in Fig.~\ref{fig:sample}, where the image (c) with pleasant texture is disfavored by the existing FR-IQA models, due to the large distance caused by the generated textures. Additionally, as shown in the right case of Fig.~\ref{fig:sample}, human perception 
is often invariant to the texture resampling, even though such resampling may lead to distance amplification in both signal and feature spaces.

In light of the aforementioned limitations, a more robust FR-IQA model is needed to address these challenges. In this paper, we propose a Deep Distance Correlation~(DeepDC) FR-IQA model, aiming to achieve consistent agreement with human perception even for the imperceptible perturbation scenarios. Specifically, the distance correlation is computed as the \red{ratio} of their distance covariance to the product of their distance standard deviations~\cite{szekely2007measuring,szekely2023energy}. For the discrete observations (\ie, deep features), the sample distance covariance has a closed-form solution that computes the inner product of the double-centered distance matrixes~\cite{dokmanic2015euclidean,szekely2009brownian}. Distance correlation is capable of capturing both linear and nonlinear relationships, providing a more holistic representation of texture information. Moreover, DeepDC does not require point-by-point alignment between reference and distorted images, which could make it more robust to imperceptible transformations and misalignments. As such, by providing a more comprehensive representation in the feature space, DeepDC aligns better with human judgments of perceptual quality~(see Fig.~\ref{fig:sample}) and generates visually appealing texture image~(see Fig.~\ref{fig:sample_texture}~(e)). In summary, the main contributions of DeepDC are threefold.

\begin{itemize}
    \item We develop a novel FR-IQA model that fully utilizes the \textit{texture-sensitive} of pre-trained DNN features, which computes distance correlation in the deep feature domain. It is worth noting that the proposed model is exclusively based on the features of the pre-trained DNNs and does not rely on fine-tuning with MOSs. 
    \item We reveal that the measurement of both linear and nonlinear relationships of the pre-trained features provides a more complete view of the deep representation, exhibiting notable flexibility and generalization capabilities. Extensive experiments achieve superior performance on five standard IQA datasets, one perceptual similarity dataset, two texture similarity datasets, and one geometric transformation dataset.
    \item We instantiate the DeepDC as an objective function for the optimization of two additional image processing tasks: texture synthesis and neural style transfer. Both quantitative and qualitative experimental results indicate that our proposed approach generates visually appealing images as well as exhibits strong generalizability, making it a potential tool for other tasks.
\end{itemize}

\section{Related Works}
\bl{ 
In this section, we provide a comprehensive review of the FR-IQA model, categorizing it into knowledge-driven and data-driven models based on their underlying mechanisms and sources of predictive power~\cite{duanmu2021quantifying}. Knowledge-driven models primarily leverage established theoretical principles, while data-driven models learn from extensive datasets.
}

\subsection{Knowledge-Driven FR-IQA Models}
For over half a century, knowledge-driven approaches have been the predominant methodology employed in FR-IQA models~\cite{mantiuk2011hdr,xue2013gradient,fang2017objective,min2017unified,wang2016just}. The pioneer FR-IQA works capture the image distortion relying on the deterministic comparisons~\cite{mannos1974effects}, such as the MSE and peak signal-to-noise ratio (PSNR). They directly compute the differences in the pixel space, which enjoys calculation simplicity and mathematical convenience. However, such methods show poor correlation with the human visual system~(HVS)~\cite{lin2003discriminative,wang2009mean}. Thus, a family of error visibility methods had been developed to find more perceptual meaningful spaces that are suited for the MSE operation, such as discrete cosine transform~\cite{watson1993dctune,malo1997subjective}, steerable pyramid and contrast normalization~\cite{teo1994perceptual}, discrete wavelet transform~\cite{watson1997visibility}, contrast sensitive function~\cite{larson2010most}, normalized Laplace pyramid~\cite{laparra2016perceptual}. Subsequently, Wang~\etal proposed to measure the visual fidelity using the structure similarity since HVS are highly sensitive to the structure information~\cite{wang2004image}. SSIM was extended to several more advanced quality measures, including multi-scale SSIM (MS-SSIM)~\cite{wang2003multiscale}, information weighting SSIM (IW-SSIM)~\cite{wang2010information}, complex wavelet SSIM (CW-SSIM)~\cite{wang2005translation}. \bl{Zhang ~\etal proposed the feature similarity (FSIM)~\cite{zhang2011fsim} and visual saliency similarity (VSI)~\cite{zhang2014vsi} for better quality assessment.} Besides, information theoretic models assume reference and distorted images are sampled from one probability distribution. The divergence between two probability distributions is deemed as the visual quality. Generally speaking, the statistical and perceptual image modeling of visual signals is the dual problem~\cite{sheikh2005information}. The information fidelity criterion (IFC) used the Gaussian scale mixture to model the natural scene statistic (NSS)~\cite{sheikh2005information}. To mimic the distortion channel of the HVS, visual information fidelity (VIF) mapped the image to the wavelet domain and calculated
the mutual information for information loss~\cite{sheikh2006image}. Other statistical models~\cite{chang2013sparse,wang2006quality} and divergence criteria~\cite{soundararajan2011rred,lin2016cross} \bl{have} been utilized to model the NSS. Moreover, fusion-based IQA models \bl{leverage} the diversity and complementarity of current methods to formulate more completeness quality evaluators, thereby enhancing the predictive performance of quality assessment~\cite{liu2012image}. Although knowledge-driven FR-IQA models have exhibited remarkable performance in visual quality prediction, they struggle to cope with imperceptible transformations such as mild geometric alterations and changes in texture perception. In this work, we propose utilizing distance correlation to construct a statistical model that can better tolerate such imperceptible transformations, as shown in Fig.~\ref{fig:sample}.

\subsection{Data-Driven FR-IQA Models}
In the deep-learning era, a great number of works concentrate on image comparison in the DNN pre-trained feature space. Gao~\etal proposed to measure the feature similarity from the VGG~\cite{simonyan2014very} and many pooling strategies were investigated to obtain the final quality score~\cite{gao2017deepsim}. In LPIPS~\cite{zhang2018unreasonable}, the multi-scale features were extracted from the pre-trained DNN networks, and the image quality was estimated by measuring the feature fidelity loss with Euclidean distance. Ghildyal~\etal enabled the LPIPS to a shift-tolerant metric~(ST-LPIPS) by studying the pooling layer, striding, padding, and anti-aliasing filtering~\cite{ghildyal2022shift}. In addition, the combination of spatial averages and correlations of the feature maps was adopted in DISTS~\cite{ding2020image}, aiming for the estimation of texture similarity and structure similarity.  The work was further improved by processing the structure and texture information adaptively~\cite{ADISTS}. Instead of measuring the feature distance point-by-point, the Wasserstein distance was utilized in DeepWSD~\cite{liao2022deepwsd} to capture the quality contamination. Driven by the quality annotated data, the human perception knowledge can also be learned by DNN from scratch, such as deep quality assessment (DeepQA)~\cite{kim2017deep}, weighted average deep image quality measure for FR-IQA (WaDIQaM-FR)~\cite{bosse2017deep}, PieAPP~\cite{prashnani2018pieapp}, perceptual information metric (PIM)~\cite{bhardwaj2020unsupervised} and joint semi-supervised and positive-unlabeled learning (JSPL)~\cite{cao2022incorporating}. However, such models rely on large-scale IQA datasets for training, which not only constrains the generalization capability of the model but also suffers from the over-fitting problem due to the limited labeled data. 

\section{DeepDC: Deep Distance Correlation}
In this section, we will give a detailed introduction to the distance correlation. Next, we instantiate the distance correlation to the pre-trained DNN~\cite{simonyan2014very}, formulating the DeepDC FR-IQA model. 

\subsection{Preliminaries of Distance Correlation}
The \textit{distance correlation}~\cite{szekely2007measuring,szekely2009brownian} was proposed to measure the dependence between two random vectors $X \in \mathbb{R}^p$ and $Y \in \mathbb{R}^q$ where $p$ and $q$ represent their vector dimensions, respectively. The distance correlation $\mathcal{V}(X, Y)=0$ indicates $X$ and $Y$ are independence. Assuming $X$ and $Y$ have finite first moments, analogous to classical covariance, a measure of dependence $\mathcal{V}^2(X, Y; w)$ is defined as follows,
\begin{equation}
\begin{aligned}
\mathcal{V}^2(X, Y ; w) &=\left\|f_{X, Y}(t, s)-f_X(t) f_Y(s)\right\|_w^2,
\end{aligned}
\end{equation}
 where $f_{XY}$ is the joint characteristic function of $X$ and $Y$ and $f_{X}$ and $f_{Y}$ denote the corresponding marginal characteristic functions. The $\|\cdot\|_w^2$-norm  represents the weighted $L_2$-norm space of functions on $\mathbb{R}^{p+q}$, which is defined as follows,
\begin{equation}
\|\gamma(t, s)\|_w^2=\int_{\mathbb{R}^{p+q}}|\gamma(t, s)|^2 w(t, s) d t  d s.
\end{equation}
The $w(t, s)$ is a positive weight function for which the integral above exists. As such,
\begin{equation}
\begin{aligned}
\mathcal{V}^2(X, Y ; w) =\int_{\mathbb{R}^{p+q}}\left|f_{X, Y}(t, s)-f_X(t) f_Y(s)\right|^2 w(t, s) d t d s.
\label{v1}
\end{aligned}
\end{equation}
To endow the $\mathcal{V}^2(X, Y; w)$ with the capability to capture  the dependence  between  $X$ and $Y$, a suitable  weight function can be found as follows \cite{szekely2009brownian}, 
\begin{equation}
w(t, s)=\left(c_p c_q|t|_p^{1+p}|s|_q^{1+q}\right)^{-1},
\label{wts}
\end{equation}
where
\begin{equation}
c_p=\frac{\pi^{(1+p) / 2}}{\Gamma((1+p) / 2)},  \quad  c_q=\frac{\pi^{(1+q) / 2}}{\Gamma((1+q) / 2)}, 
\end{equation}
and $\Gamma (\cdot)$ is the complete gamma function~\cite{abramowitz1988handbook}. Based on Eqn.~(\ref{v1}) and Eqn.~(\ref{wts}),  the measures of distance covariance and correlation can be defined as follows.
\begin{definition}[Distance covariance~\cite{szekely2007measuring}]
\label{def:dcov}
The distance covariance between random vector $X$ and $Y$ is the square root of $\mathcal{V}^2(X,Y)$ defined by

\begin{align}
\mathcal{V}^{2}(X, Y) &=\left\|f_{X, Y}(t, s)-f_{X}(t) f_{Y}(s)\right\|^{2} \\
&=\frac{1}{c_{p} c_{q}} \int_{\mathbb{R}^{p+q}} \frac{\left|f_{X, Y}(t, s)-f_{X}(t) f_{Y}(s)\right|^{2}}{|t|_{p}^{1+p}|s|_{q}^{1+q}} dt ds.
\end{align}
Likewise, the \textit{distance variance}  is the square root of $\mathcal{V}^2(X,X)=\mathcal{V}^2(X)=\left\|f_{X, X}(t, s)-f_{X}(t) f_{X}(s)\right\|^{2}$.
\end{definition}
\begin{definition}[Distance correlation~\cite{szekely2007measuring}]
\label{def:dCor} The \textit{distance correlation} between random vector $X$ and $Y$ is the square root of $\mathcal{R}^2(X,Y)$, which is defined by the division of distance covariance to the product of corresponding distance variance
\begin{align}
    \mathcal{R}^2(X, Y)= \left\{\begin{matrix}
 \frac{\mathcal{V}^2(X,Y)}{\sqrt{\mathcal{V}^2(X)\mathcal{V}^2(Y)}}, & \mathcal{V}^2(X)\mathcal{V}^2(Y)>0,   \\ 
0, & \mathcal{V}^2(X)\mathcal{V}^2(Y)=0.
\end{matrix}\right.
\end{align}

\end{definition}
In practice, the observations of $X$ and $Y$ are usually discrete. Therefore, sampling $n$ observations $(\mathbf{X}, \mathbf{Y}) =\left\{\left(X_k, Y_k\right): k=1, \ldots, n\right\}$ from the the joint distribution of random vectors $X \in \mathbb{R}^p$ and $Y \in \mathbb{R}^q$. The sample distance covariance has a simple computing formula, which can be defined as follows.
\begin{definition}[Sample distance covariance~\cite{szekely2007measuring}] \label{def:sdCor}
The emperical sample distance covariance between sample vector $\mathbf{X}$ and $\mathbf{Y}$ is define by
\begin{align}\label{pro}
\mathcal{V}^{2}_n(\mathbf{X}, \mathbf{Y}) = \frac{1}{n^{2}} \sum_{k, l = 1}^{n}{\bm{A}}_{k l} {\bm{B}}_{k l},
\end{align}
where the ${\bm{A}}_{k l}$ and ${\bm{B}}_{k l}$ are double-centered distance matrixes of $\mathbf{X}$ and $\mathbf{Y}$, respectively. The ${\bm{A}}_{k l}=a_{k l}-\bar{a}_{k \cdot}-\bar{a}_{\cdot l}+\bar{a}_{. .}$ and 
\begin{equation}
\begin{aligned}
a_{k l}=\left\|X_k-X_l\right\|_z, \quad \bar{a}_{k \cdot}=\frac{1}{n} \sum_{l=1}^n a_{k l}, \quad \\ 
\bar{a}_{\cdot l},=\frac{1}{n} \sum_{k=1}^n a_{k l},  \quad  \bar{a}_{. .} =\frac{1}{n^2} \sum_{k, l=1}^n a_{k l},
\end{aligned}
\label{akl}
\end{equation}
where $\|\cdot\|_z$ means the z-norm.  Analogously, we define ${\bm{B}}_{k l}=b_{k l}-\bar{b}_{k \cdot}-\bar{b}_{\cdot l}+\bar{b}_{. .}$, for $k,l =1, \ldots, n$. The sample distance variance is defined by $\mathcal{V}^2_n(\mathbf X, \mathbf X)=\mathcal{V}^2_n(\mathbf X)= \frac{1}{n^{2}} \sum_{k, l = 1}^{n}{\bm{A}}_{k l}^2$.
\end{definition}
The proof of the sample distance covariance based on Definition~\ref{def:dcov} equivalence with Definition~\ref{def:sdCor} can be found in~\cite{szekely2023energy}.

\begin{figure}
    \centering
    \captionsetup{justification=centering}
    \subfloat{\includegraphics[trim={0cm 0cm 0cm .2cm},clip,width=.23\textwidth]{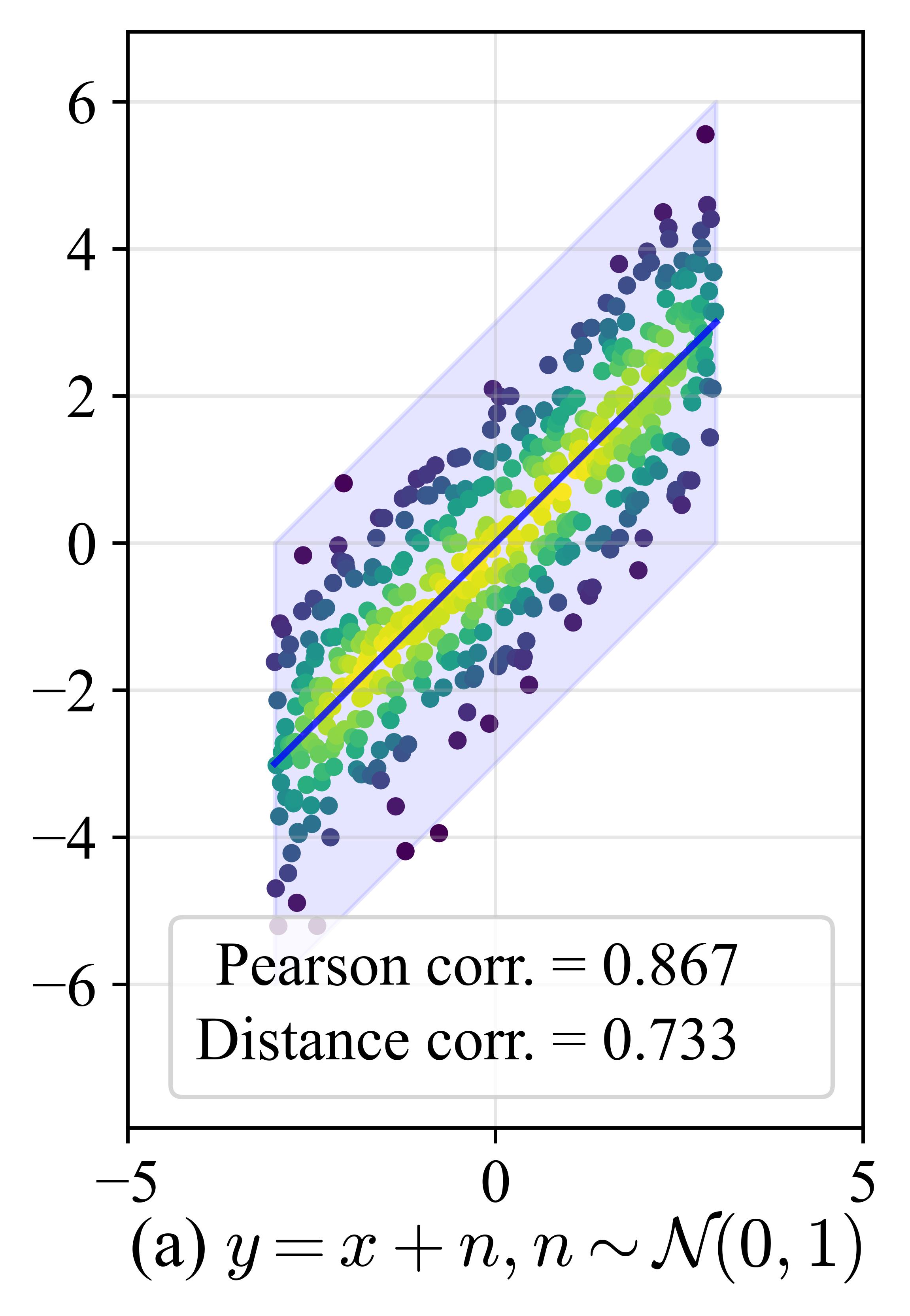}}\hskip.1em
    \subfloat{\includegraphics[trim={0cm 0cm 0cm .2cm},clip,width=.242\textwidth]{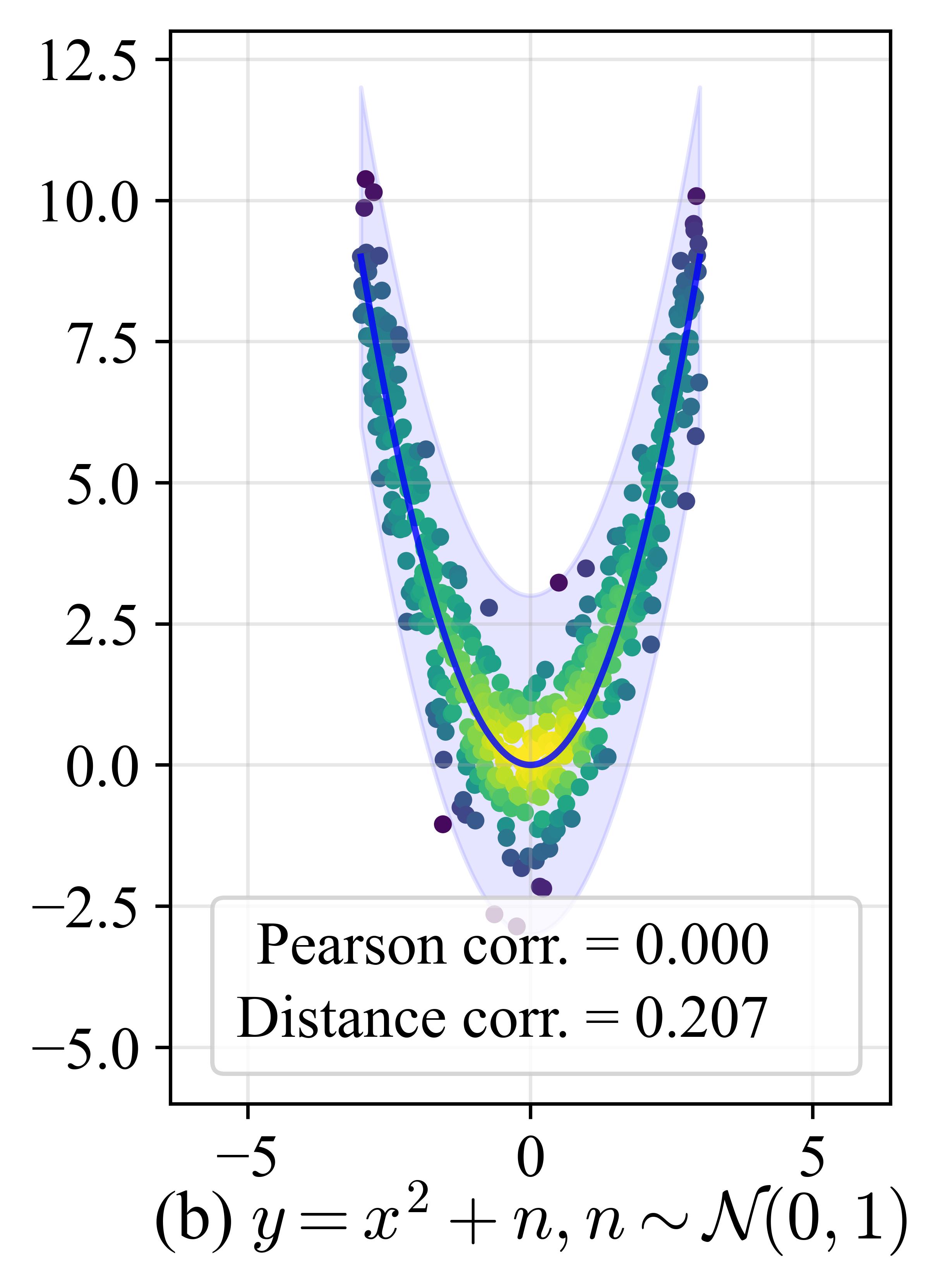}}\\
    \caption{\red{The illustration of linear and nonlinear relationships through the Pearson correlation and distance correlation. (a) Linear relationship; (b) Nonlinear relationship. }}\label{fig:toy}
\end{figure}

\begin{definition}[Sample distance correlation~\cite{szekely2007measuring}]
The empirical sample distance correlation between sample $\mathbf{X}$ and $\mathbf{Y}$ is defined by
\begin{align}\label{def:sdc}
    \mathcal{R}^2_n(\mathbf X, \mathbf Y)= \left\{\begin{matrix}
 \frac{\mathcal{V}^2_n(\mathbf X,\mathbf Y)}{\sqrt{\mathcal{V}^2_n(\mathbf X)\mathcal{V}^2_n(\mathbf Y)}}, & \mathcal{V}^2_n(\mathbf X)\mathcal{V}^2_n(\mathbf Y)>0,   \\ 
0, & \mathcal{V}^2_n(\mathbf X)\mathcal{V}^2_n(\mathbf Y)=0.
\end{matrix}\right.
\end{align}
\end{definition}
As such, distance correlation is both straightforward to compute and effective at identifying nonlinear relationships. In contrast to Pearson correlation, 2D toy examples utilizing distance correlation are presented in Fig.~\ref{fig:toy}. These examples demonstrate that distance correlation is capable of measuring both linear and nonlinear relationships concurrently. When the relationship between two random variables is not linearly correlated, the Pearson correlation is close to zero but the distance correlation remains significant.

\subsection{Feature Transformation}
Given the reference image ${\bm {x}  \in \mathbb{R}^{h \times w \times 3}}$ and the \bl{distorted} image ${\bm {y}  \in \mathbb{R}^{h\times w \times 3}}$, the aim of FR-IQA is to predict the image quality score of the distorted image. Herein, instead of directly calculating the dependency between ${\bm x}$ and  ${\bm y}$ in the pixel space, we chose to transform the image to the deep feature domain as human perceptual sensitivity is usually non-uniform~\cite{wang2008maximum,berardino2017eigen}.  
More importantly, the intrinsic \textit{texture-sensitive} characteristic of the pre-trained DNNs plays a crucial role in building effective perceptual quality assessment models~\cite{zhang2018unreasonable,prashnani2018pieapp,ding2020image}.  
Following the vein, we also adopt the VGG19 network~\cite{simonyan2014very} to nonlinearly transform the images (${\bm x}$ and  ${\bm y}$) to the deep representations ($\tilde{\bm x}$ and  $\tilde{\bm y}$). In addition, the fine-grained representations are fit for the multi-scale processing mechanism of the HVS and have demonstrated great success in quality assessment tasks~\cite{simoncelli1995steerable,wang2003multiscale,zhu2022learing}. As such, we include the multi-scale features from the five stages of the convolutional layers, which can be expressed as follows:
\begin{equation}
\begin{aligned}
    \tilde{\bm  x}_i^j= f_i^j(\bm x),\, i=0, \dots, n^j, j=1, \dots, m,\\
    \tilde{\bm  y}_i^j= f_i^j(\bm y),\, i=0, \dots, n^j, j=1, \dots, m,
\end{aligned}  
\end{equation}
where $f_i^j(\cdot)$ indicates the feature transform $f$ is performed at the $j$-th layer of the $i$-th feature map. The total number of layers $m$ is equal to five, denoted as \texttt{conv1\_2}, \texttt{conv2\_2}, \texttt{conv3\_3}, \texttt{conv4\_3}, \texttt{conv5\_3}, respectively. The $n^j$ represents the number of feature maps in the $j$-th layer. $\tilde{\bm x}_i^j \in \mathbb{R}^{h^j\times w^j \times n^j}$ and $\tilde{\bm y}_i^j \in \mathbb{R}^{h^j\times w^j \times n^j}$ represent the corresponding feature maps, and $h^j$ and $w^j$ are the spatial height and width of the feature map in the $j$-th layer.

\begin{figure}[t]
 \centering
 \includegraphics[trim={1cm 0cm 0cm 0cm},clip,width=.49\textwidth]{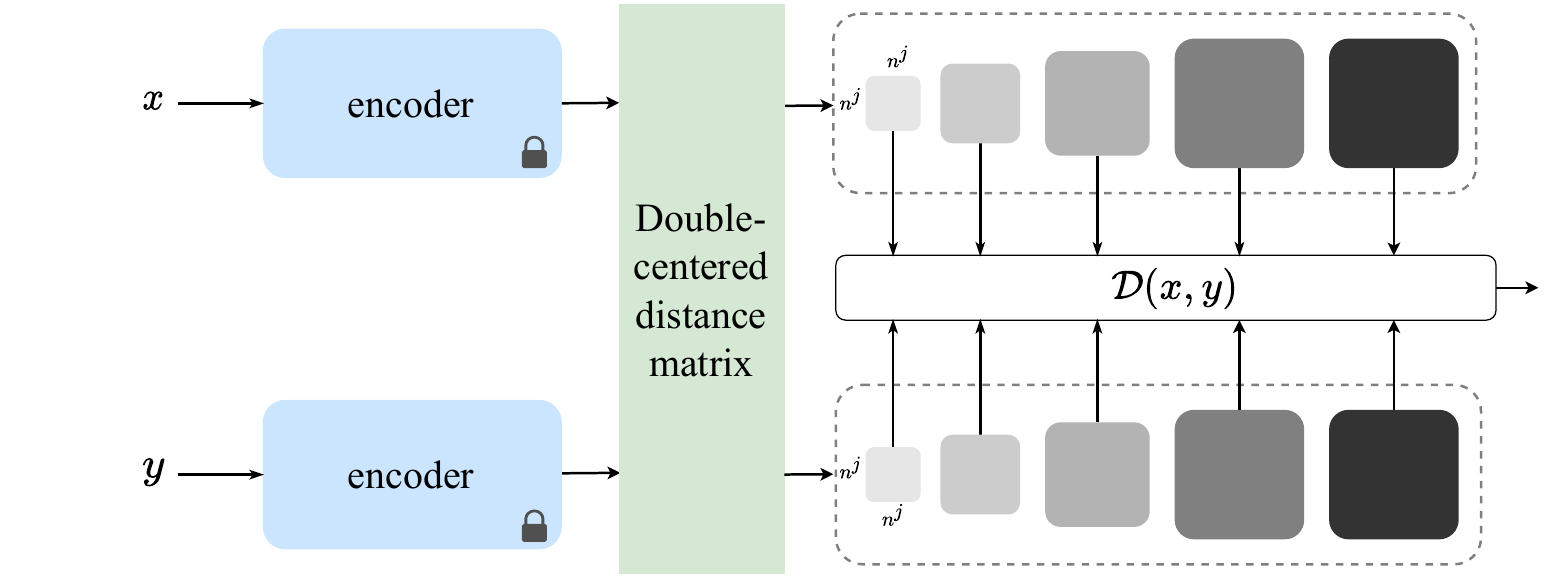}
 \caption{Overall structure of the proposed model. The pre-trained VGG19 network is adopted as the deep feature extractor. Then, the multi-scale features from five convolutional layers (\texttt{conv1\_2}, \texttt{conv2\_2}, \texttt{conv3\_4}, \texttt{conv4\_4}, \texttt{conv5\_4}) are fed to compute the corresponding double-centered distance matrix using Eqn.~(\ref{eq:edm}). The final quality score is obtained by the arithmetic average of the distance correlation across five convolutional layers (referring to Eqn.~(\ref{eq:quality})).}
 \label{fig:frame}
\end{figure}

\subsection{Perceptual Quality Measure}
With the obtained multi-scale features ($\tilde{\bm x}$ and  $\tilde{\bm y}$), we apply the distance correlation (Eqn.~(\ref{def:sdc})) to measure the statistical similarity, formulating the DeepDC FR-IQA measure. The overall design framework of the proposed model is shown in Fig.~\ref{fig:frame}. Particularly, we first calculate the distance covariance at each convolutional layer using Eqn.~(\ref{def:sdCor}), which can be formulated as follows,
\begin{align}
    \mathcal{V}^{2}_{n^j}(\tilde{\bm x}_i^j,  \tilde{\bm y}_i^j)=\frac{1}{(n^j)^2}\sum_{i,k=1}^{n^j}\bm{A}_{ik}^j\bm{B}_{ik}^j,\,
\end{align}
where $\mathcal{V}^{2}_{n^j}(\tilde{\bm x}_i^j,  \tilde{\bm y}_i^j) \in \mathcal{R}^{n^j \times n^j}$ indicates the distance covariance of $\tilde{\bm x}_i^j$ and  $\tilde{\bm y}_i^j$ in the $j$-th convolutional layer with $n^j$ feature maps (discrete observations). ${\bm{A}}_{ik}^j \in \mathcal{R}^{n^j \times n^j}$ and ${\bm{B}}_{ik}^j \in \mathcal{R}^{n^j \times n^j}$ stand for the double-centered distance matrixes, which is computed by ${\bm{A}}_{ik}^j=a_{ik}^j-\bar{a}_{i \cdot}^j-\bar{a}_{\cdot k}^j+\bar{a}_{. .}^j$ and ${\bm{B}}_{i k}^j=b_{i k}^j-\bar{b}_{i \cdot}^j-\bar{b}_{\cdot k}^j+\bar{b}_{. .}^j$. Taking ${\bm{A}}_{ik}^j$ as an instance, we adopt $z=2$ in Eqn.~(\ref{akl}), 
\begin{equation}
\begin{aligned}\label{eq:edm}
a_{ik}^j=\left\|\tilde{\bm x}_i^j-\tilde{\bm x}_k^j\right\|_2, \quad 
\bar{a}_{i \cdot}^j=\frac{1}{n^j} \sum_{l=1}^{n^j} a_{i k}^j, \quad \\ 
\bar{a}_{\cdot k}^j,=\frac{1}{n^j} \sum_{i=1}^{n^j} a_{i k}^j,  \quad  
\bar{a}_{. .}^j =\frac{1}{(n^j)^2} \sum_{i, k=1}^n a_{i k}^j.
\end{aligned}
\end{equation}
\red{The double-centered distance matrix is a new texture statistic capable of measuring both linear and nonlinear relationships of the deep features. It facilitates the model to fully capture the texture information in the pre-trained features, which are validated in Sec.~\ref{sec:vis} and Sec.~\ref{sec:app}.} Similarily, the distance variance in $j$-th convolutional layer with $n^j$ feature maps can be computed by $\mathcal{V}^2_{n^j}(\tilde{\bm x}_i^j, \tilde{\bm x}_i^j)=\mathcal{V}^2_{n^j}(\tilde{\bm x}_i^j)= \frac{1}{{(n^j)^2}} \sum_{i,k = 1}^{n^j}({\bm{A}}_{ik}^j)^2$.

\begin{table*}[htbp]
  \centering
  \small
  \caption{Performance comparison of DeepDC against twelve existing FR-IQA models on five standard IQA datasets in terms of SRCC and PLCC. The best two results are highlighted in boldface and underline, respectively.}
  % Table generated by Excel2LaTeX from sheet 'table_1'
    \begin{tabular}{lcccccccccccccc}
    \toprule
    \multirow{2}[4]{*}{Method} & \multicolumn{2}{c}{LIVE~\cite{sheikh2003image}} &              & \multicolumn{2}{c}{CSIQ~\cite{larson2010most}} &              & \multicolumn{2}{c}{TID2013~\cite{ponomarenko2015image}} &              & \multicolumn{2}{c}{KADID-10k~\cite{2019KADID}} &              & \multicolumn{2}{c}{PIPAL~\cite{jinjin2020pipal}} \\
    \cmidrule{2-3}\cmidrule{5-6}\cmidrule{8-9}\cmidrule{11-12}\cmidrule{14-15}             & SRCC         & PLCC         &              & SRCC         & PLCC         &              & SRCC         & PLCC         &              & SRCC         & PLCC         &              & SRCC         & PLCC \\
    \midrule
    PSNR         & 0.873        & 0.868        &              & 0.809        & 0.815        &              & 0.688        & 0.679        &              & 0.676        & 0.680        &              & 0.407        & 0.415  \\
    SSIM~\cite{wang2004image} & 0.931        & 0.928        &              & 0.872        & 0.868        &              & 0.720        & 0.745        &              & 0.724        & 0.723        &              & 0.498        & 0.505  \\
    MS-SSIM~\cite{wang2003multiscale} & 0.931        & 0.931        &              & 0.908        & 0.896        &              & 0.798        & 0.810        &              & 0.802        & 0.801        &              & 0.552        & 0.590  \\
    VIF~\cite{sheikh2006image} & 0.927        & 0.925        &              & 0.902        & 0.887        &              & 0.690        & 0.732        &              & 0.593        & 0.602        &              & 0.443        & 0.468  \\
    FSIM~\cite{zhang2011fsim} & \textbf{0.965 } & \textbf{0.961 } &              & 0.931        & 0.919        &              & \underline{0.851}        & \underline{0.877}        &              & 0.854        & 0.851        &              & 0.589        & 0.615  \\
    NLPD~\cite{laparra2016perceptual} & 0.914        & 0.914        &              & 0.917        & 0.911        &              & 0.808        & 0.823        &              & 0.810        & 0.810        &              & 0.469        & 0.509  \\
    \midrule
    PieAPP~\cite{prashnani2018pieapp} & 0.908        & 0.919        &              & 0.877        & 0.892        &              & 0.850        & 0.848        &              & 0.836        & 0.836        &              & \textbf{0.700 } & \textbf{0.712 } \\
    LPIPS~\cite{zhang2018unreasonable} & 0.939        & 0.945        &              & 0.883        & 0.906        &              & 0.695        & 0.759        &              & 0.720        & 0.729        &              & 0.573        & 0.618  \\
    DSD~\cite{kligvasser2021deep} & 0.577        & 0.552        &              & 0.603        & 0.700        &              & 0.548        & 0.657        &              & 0.439        & 0.527        &              & 0.274        & 0.350  \\
    DISTS~\cite{ding2020image} & \underline{0.955}   & \underline{0.954}        &              & 0.939        & 0.941        &              & 0.830        & 0.856        &              & -        & -        &              & 0.624        & 0.644  \\
    DeepWSD~\cite{liao2022deepwsd} & 0.952        & 0.949        &              & \textbf{0.954 } & \underline{0.947}        &              & \textbf{0.869 } & \textbf{0.893 } &              & \underline{0.881}        & \underline{0.877}        &              & 0.514        & 0.517  \\
    ST-LPIPS~\cite{ghildyal2022shift} & 0.870        & 0.868        &              & 0.861        & 0.842        &              & 0.640        & 0.678        &              & 0.759        & 0.762        &              & 0.593        & 0.599  \\
    \midrule
    DeepDC~(\textit{Ours}) & 0.951        & 0.947        &              & \underline{0.951}        & \textbf{0.956 } &              & 0.844        & 0.866        &              & \textbf{0.906 } & \textbf{0.900 } &              & \underline{0.684}        & \underline{0.687}  \\
    \bottomrule
    \end{tabular}%
   \label{tab:mainresult}%
\end{table*}%

Then, the distance correlation (denoted as $\mathcal{R}^2_{n^j}(\tilde{\bm x}_i^j,  \tilde{\bm y}_i^j) \in \mathbb{R}^{n^j \times n^j}$) measures the dependencies between the reference and distorted images   by computing the division of the distance covariance to the product of corresponding distance standard deviations, which can be expressed as follows,
\begin{align}
    \mathcal{R}^2_{n^j}(\tilde{\bm x}_i^j,  \tilde{\bm y}_i^j) = \frac{\mathcal{V}^{2}_{n^j}(\tilde{\bm x}_i^j,  \tilde{\bm y}_i^j) + \epsilon}{\sqrt{\mathcal{V}^{2}_{n^j}(\tilde{\bm x}_i^j)\mathcal{V}^{2}_{n^j}(\tilde{\bm y}_i^j)} + \epsilon}\,, 
\end{align}
where $\epsilon$ is a stable parameter to avoid division by zero. 

Finally, the proposed DeepDC model aggregates the distance
correlations from different convolution layers using a
arithmetic average, which can be formulated as follows,
\begin{align}\label{eq:quality}
    \mathcal{D}{(\bm x, \bm y)} = 1 - \frac{1}{m}\sum_{j=0}^{m}\mathcal{R}^2_{n^j}(\tilde{\bm x}_i^j,  \tilde{\bm y}_i^j)\,, 
\end{align}

As such, the $\mathcal{D}{(\bm x, \bm y)} \in [0, 1]$, \bl{and} a higher score of DeepDC indicates worse perceptual quality compared with the reference image.
It is worth noting that, besides utilizing the pre-trained parameters, the proposed DeepDC model does not necessitate fine-tuning with MOSs, which largely eliminates the risk of over-fitting. Moreover, the proposed DeepDC exhibits a plug-and-play nature, ensuring compatibility and flexibility with various DNN backbones, as validated in Sec.~\ref{sec:ablation}.

\subsection{Properties of DeepDC}
Herein, we delve into the mathematical properties of DeepDC, with the objective of elucidating how these properties contribute to its superiority in quality assessment tasks.

\begin{enumerate}[label=(\roman*)]
\item \textbf{Boundedness}: The $0 \leq \mathcal{D}(\bm x, \bm y) \leq 1$ ensures that the DeepDC model produces a normalized score, which is critical for consistent and interpretable evaluation across different images. 

\item \textbf{Sensitivity to Independence}: \bl{The $\mathcal{D}(\bm x, \bm y) = 1$ if and only if $\tilde{\bm x}$ and $\tilde{\bm y}$ are independent. This ensures that the model recognizes the maximal deviation in quality when two images in the deep feature domain are entirely unrelated.}

\item \textbf{Robustness to Linear Transformations}: 
\bl{The condition $\mathcal{D}(\bm x, \bm y) = 0$ ensures a linear relationship between $\bm x$ and $\bm y$ in the deep feature domain. This indicates that DeepDC is robust to transformations that do not affect perceptual quality such as scaling, rotation, or translation. }

\end{enumerate} 
The mathematical properties of DeepDC emphasize its alignment with human perception, particularly the robustness to perceptually imperceptible transformations. This alignment is essential for an FR-IQA model, as it should ideally mimic human judgment of image quality. 
The bounded score range of DeepDC ensures consistency and interpretability, and its sensitivity to independence facilitates effective comparison of related images. As detailed in Sec.~\ref{sec:exp}, we conduct comprehensive experiments to further support the effectiveness and generalizability of the proposed model.

\section{Experiments}
\label{sec:exp}
In this section, we first describe the experimental setups. Then, we conduct comprehensive experiments to verify the effectiveness of the proposed model, including standard image quality prediction, perceptual similarity measurement, texture similarity qualification, and mild geometric transformation assessment. Next, we explain the effectiveness with comprehensive feature visualizations. Finally, ablation studies are performed.

\subsection{Experimental Setups}
The default feature extractor is VGG19 which is pre-trained on the ImageNet~\cite{deng2009imagenet}. For all experiments, we compute distance correlations at convolutional layers \texttt{conv1\_2}, \texttt{conv2\_2}, \texttt{conv3\_4}, \texttt{conv4\_4}, \texttt{conv5\_4}, which contains $64$, $128$, $256$, $512$, $512$ feature maps, respectively.
In addition, inspired by the SSIM~\cite{wang2004image} and DISTS~\cite{ding2020image}, we resize the shorter side of the input images to $224$ while keeping the aspect ratio when we calculate the quality score of the distorted image. We apply Spearman's rank-order correlation coefficient (SRCC) and Pearson linear correlation coefficient (PLCC) to evaluate the monotonicity and linearity. The larger SRCC and PLCC values reflect better quality prediction results. In particular,  a five-parameter nonlinear logistic function is fitted to map the predicted scores to the same scale as MOSs when computing PLCC~\cite{video2000final}. 
\begin{align}
    \hat{\mathcal{D}} = \eta_1\left (\frac{1}{2} - \frac{1}{\exp(\eta_2({\mathcal{D}} - \eta_3))}\right )+\eta_4{\mathcal{D}}+\eta_5  ,
\end{align}
where $\{\eta_i\}_{i=1}^{5}$ are the fitting parameters. 
\begin{table}[t]
  \centering
  \small
	\caption{Performance comparison of DeepDC against state-of-the-art methods on the GAN distortion of the PIPAL dataset in terms of SRCC and PLCC. The measure specifically designed for GAN images is represented in italics.}\label{tab:gan}	
    % Table generated by Excel2LaTeX from sheet 'table_2'
    \begin{tabular}{lcc}
    \toprule
    \multirow{2}[4]{*}{Method} & \multicolumn{2}{c}{PIPAL (GAN distortions)} \\
    \cmidrule{2-3}             & SRCC         & PLCC  \\
    \midrule
    PSNR         & 0.2911       & 0.416 \\
    SSIM~\cite{wang2004image} & 0.322        & 0.472  \\
    MS-SSIM~\cite{wang2003multiscale} & 0.387        & 0.615  \\
    VIF~\cite{sheikh2006image} & 0.324        & 0.543  \\
    FSIM~\cite{zhang2011fsim} & 0.410        & 0.621  \\
    NLPD~\cite{laparra2016perceptual} & 0.341        & 0.570  \\
    \midrule
    PieAPP~\cite{prashnani2018pieapp} & \underline{0.553}        & \underline{0.632}  \\
    LPIPS~\cite{zhang2018unreasonable} & 0.486        & 0.617  \\
    DISTS~\cite{ding2020image} & 0.549        & 0.607  \\
    DeepWSD~\cite{liao2022deepwsd} & 0.397        & 0.560  \\
    ST-LPIPS~\cite{ghildyal2022shift} & 0.492        & 0.490  \\
    \midrule
    \textit{FID} & 0.413        & 0.496  \\
    \midrule
    DeepDC~(\textit{Ours})       & \textbf{0.588} & \textbf{0.638} \\
    \bottomrule
    \end{tabular}%
\end{table}%

\begin{table*}[t]
    \caption{Performance comparison on the validation set of BAPPS~\cite{zhang2018unreasonable} with 2AFC configuration. \bl{The 2AFC scores lie in $[0, 1]$, and a higher value indicates better performance.}}
    \label{tab:bapps}
    \centering
    \resizebox{\linewidth}{!}{
    \begin{tabular}{l*{9}{c}}
    \toprule
    \multirow{3}{*}{Method} & \multicolumn{3}{c}{Synthetic distortions} & \multicolumn{5}{c}{Distortions by real-world algorithms} & \multirow{3}{*}{All} \\
    \cmidrule(lr){2-4}
    \cmidrule(lr){5-9}
    & Traditional & CNN-based & All & \makecell[c]{Super \\ resolution} & \makecell[c]{Video \\ deblurring} & Colorization & \makecell[c]{Frame \\ interpolation} & All & 
    \\ \midrule
    Human                             & 0.808 & 0.844 & 0.826 & 0.734 & 0.671 & 0.688 & 0.686 & 0.695 & 0.739 \\ 
    \midrule
    PSNR                              & 0.573 & 0.801 & 0.687 & 0.642 & 0.590 & \underline{0.624} & 0.543 & 0.613 & 0.616 \\ 
    SSIM~\cite{wang2004image}         & 0.575 & 0.769 & 0.672 & 0.616 & 0.581 & 0.521 & 0.554 & 0.584 & 0.607 \\ 
    MS-SSIM~\cite{wang2003multiscale} & 0.588 & 0.767 & 0.678 & 0.638 & 0.589 & 0.523 & 0.572 & 0.596 & 0.618\\
    VIF~\cite{sheikh2006image}        & 0.556 & 0.744 & 0.650 & 0.651 & 0.594 & 0.514 & 0.597 & 0.589 & 0.615 \\
    FSIM~\cite{zhang2011fsim}         & 0.628 & 0.794 & 0.711 & 0.661 & 0.590 & 0.571 & 0.581 & 0.614 & 0.640 \\
    NLPD~\cite{laparra2016perceptual} & 0.550 & 0.766 & 0.658 & 0.634 & 0.585 & 0.526 & 0.538 & 0.591 & 0.609 \\
    \midrule
    PieAPP~\cite{prashnani2018pieapp} & 0.703 & 0.765 & 0.734 & 0.619 & 0.547 & 0.612 & 0.607 & 0.592 & 0.629 \\
    LPIPS~\cite{zhang2018unreasonable}& 0.714 & \underline{0.814} & 0.764 & 0.690 & 0.573 & 0.618 & 0.627 & 0.627 & 0.662 \\
    DSD~\cite{kligvasser2021deep}     & 0.580 & 0.698 & 0.639 & 0.654 & 0.560 & 0.515 & 0.595 & 0.591 & 0.605 \\ 
    DISTS~\cite{ding2020image}        & \underline{0.749} & 0.813 & \textbf{0.797} & \underline{0.692} & \underline{0.599} & 0.614 & \underline{0.627} & \underline{0.642} & \underline{0.678} \\
    DeepWSD~\cite{liao2022deepwsd}    & 0.594 & 0.788 & 0.691 & 0.630 & 0.588 & 0.569 & 0.547 & 0.597 & 0.623 \\ 
    % ST-LPIPS~\cite{ghildyal2022shift} & 0.500 & 0.500 & 0.500 & 0.500 & 0.500 & 0.501 & 0.500 & 0.500 & 0.500 \\
    \midrule
    DeepDC~(\textit{Ours})           & \textbf{0.757} & \textbf{0.815} & \underline{0.785} & \textbf{0.704} & \textbf{0.615} & \textbf{0.629} & \textbf{0.631 } & \textbf{0.650} & \textbf{0.685} \\
    \bottomrule
    \end{tabular}
    }
\end{table*}

\subsection{Performance on Image Quality Prediction}
We compare DeepDC with $12$ FR-IQA models on five standard IQA datasets, including the LIVE~\cite{sheikh2003image} CSIQ~\cite{larson2010most}, TID2013~\cite{ponomarenko2015image}, KADID-10k~\cite{2019KADID}, and PIPAL~\cite{jinjin2020pipal}. In particular, LIVE, CSIQ, and TID2013 contain limited image contents and distortion types, and they have been widely used for more than ten years. The KADID-10k and PIPAL are two large-scale IQA datasets with more than ten thousand distorted images. KADID-10k has $81$ pristine images, and $25$ distortion types with $5$ levels are adopted to generate $10,125$ distorted images. PIPAL is one of the largest human-rated IQA datasets with $23,200$ images, which are generated by $200$ reference images with $40$ distortion types. It is worth noting that PIPAL introduces $19$ GAN-based distortions, challenging the existing FR-IQA models a lot. In addition, the $12$ FR-IQA models cover various design methodologies: the error visibility methods - PSNR, and NLPD~\cite{laparra2016perceptual}, the structural similarity methods - SSIM~\cite{wang2004image}, MS-SSIM~\cite{wang2003multiscale}, and FSIM~\cite{zhang2011fsim}; the information-theoretic methods - VIF~\cite{sheikh2006image}; the learning-based methods - PieAPP~\cite{prashnani2018pieapp}, LPIPS~\cite{zhang2018unreasonable}, and ST-LPIPS~\cite{ghildyal2022shift}; the distribution-based methods - DISTS~\cite{ding2020image}, DSD~\cite{kligvasser2021deep}, DeepWSD~\cite{liao2022deepwsd}. Notably, we use the PyTorch implementations\footnote{\url{https://github.com/dingkeyan93/IQA-optimization/tree/master/IQA_pytorch}} to reproduce the results of SSIM, MS-SSIM, FSIM, VIF, and NLPD and the remaining models calculate the quality scores using their original implementations. \bl{Additionally, DISIS was trained on the KADID-10k dataset such that the results on KADID-10k are not reported}. The entire comparative experiment is conducted utilizing the VGG configuration for the LPIPS and ST-LPIPS.

The IQA experimental results are listed in Table~\ref{tab:mainresult}, from which we can find DeepDC achieves superior performance on both classical~(LIVE, CSIQ, and TID2013) and latest~(KADID-10k and PIPAL) IQA datasets.
It demonstrates that DeepDC achieves competitive performance on all datasets, and the independence from MOS endows the DeepDC with the roust generalization capability, especially on the large-scale IQA datasets~\cite{2019KADID,jinjin2020pipal}.
In addition, the deep learning-based methods perform worse than the knowledge-driven methods~(\eg, FSIM) on small-scale IQA datasets, indicating the potential over-fitting problem. 
Moreover, DeepWSD outperforms most learning-based methods on synthetic distortions, which further reflects the success of the advanced distribution comparison for building the FR-IQA model. 
Finally, though PieAPP obtains the best performance on the PIPAL dataset, it requires plenty of human-rated images to train the model~\cite{prashnani2018pieapp}.

Particularly, we further compare DeepDC with the FR-IQA models on the GAN-generated images of the PIPAL dataset. As shown in Table~\ref{tab:gan}, most distance-based FR-IQA models present poor performance, and the underlying reason may lie in that synthesized textures which not appear in reference images are usually introduced by the generation networks.  Although the FID~\cite{heusel2017gans} is designed especially for the quality evaluation of GAN models, the poor performance reveals its limitations and the great challenges of GAN-generated IQA.  Compared with those methods, our model achieves the best result, which indicates the distribution rather than the distance mitigates the strict requirement of point-by-point alignment during feature comparison.

\begin{table}[t]
  \centering
  \small
	\caption{Performance comparison of DeepDC against state-of-the-art methods on two texture similarity datasets in terms of SRCC and PLCC. The texture similarity models are represented in italics.}\label{tab:texutre_similarity}	
    % Table generated by Excel2LaTeX from sheet 'table_2'
    \begin{tabular}{lccccc}
    \toprule
    \multirow{2}[4]{*}{Method } & \multicolumn{2}{c}{SynTEX~\cite{golestaneh2015effect}} && \multicolumn{2}{c}{TQD~\cite{ding2020image}} \\
    \cmidrule{2-3} \cmidrule{5-6}             & SRCC         & PLCC  &       & SRCC         & PLCC \\
    \midrule
    SSIM~\cite{wang2004image} & 0.579        & 0.598   &     & 0.352        & 0.418  \\
    VIF~\cite{sheikh2006image} & 0.606        & 0.697   &     & 0.549        & 0.614  \\
    FSIM~\cite{zhang2011fsim} & 0.081        & 0.115   &     & 0.386        & 0.272  \\
    NLPD~\cite{laparra2016perceptual} & 0.606        & 0.607   &    & 0.409        & 0.457  \\
    \midrule
    PieAPP~\cite{prashnani2018pieapp} & 0.715        & 0.719   &     & 0.718        & 0.721  \\
    LPIPS~\cite{zhang2018unreasonable} & 0.788        & 0.788   &     & 0.203        & 0.188  \\
    DISTS~\cite{ding2020image} & \textbf{0.922 } & \textbf{0.894 } & &\underline{0.874}        & 0.863  \\
    % WSD~\cite{kligvasser2021deep} & 0.853        & 0.877      &  & 0.852        & \underline{0.880}  \\
    ST-LPIPS~\cite{ghildyal2022shift} & 0.341        & 0.397  &      & 0.400        & 0.428  \\
    \midrule
    \textit{STSIM~\cite{zujovic2013structural}} & 0.643        & 0.650 &       & 0.408        & 0.422  \\
    \textit{NPTSM~\cite{alfarraj2016content}} & 0.496        & 0.505   &     & 0.679        & 0.678  \\
    \textit{IGSTOA~\cite{golestaneh2018synthesized}} & 0.820        & 0.816    &    & 0.802        & 0.804  \\
    \midrule
    DeepDC (\textit{Ours}) & \underline{0.901}        & \underline{0.881}    &    & \textbf{0.889 } & \textbf{0.914 } \\
    \bottomrule
    \end{tabular}%
    
\end{table}%

\subsection{Performance on Perceptual Similarity Measurement}
Herein, we also test the perceptual similarity of the proposed and comparison models on the validation set of the BAPPS~\cite{zhang2018unreasonable}. BAPPS contains a great number of image patches, which have been degraded by synthetic distortion~(\eg, noise, blurring, and compression) or generated by real-world algorithms such as super-resolution and colorization. Human judgments on perceptual similarity in BAPPS are collected using a two-alternative forced choice~(2AFC) experimental setup. The 2AFC score is introduced as a criterion, computing based on the consistency of the model's preference with human judgment. The results of the competing methods are calculated with the shorter side of the image patches adjusted to 256.  The results are listed in Table \ref{tab:bapps}, from which we can find the proposed model demonstrates superior performance on both synthetic and real-world distortions, especially excelling in handling distortions created by real-world algorithms. It outperforms other methods with the highest score of $0.650$, which is close to human judgment~($0.695$). Additionally, though deep learning-based models~(\eg, PieAPP, LPIPS, and DISTS) are trained on large-scale IQA datasets, they still underperform the proposed model.

\begin{figure*}
    \centering
    \captionsetup{justification=centering}
    \subfloat[AWGN]{\includegraphics[scale=0.5,width=0.16\textwidth]{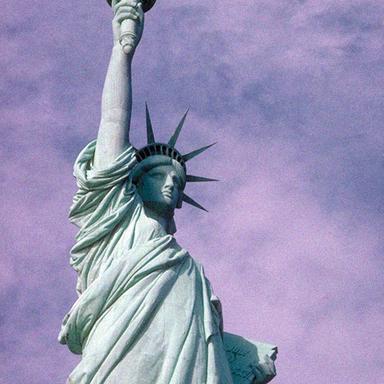}}\hskip.2em
    \subfloat[GBlur]{\includegraphics[scale=0.5,width=0.16\textwidth]{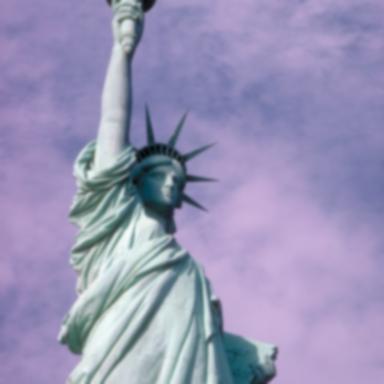}}\hskip.2em
    \subfloat[Contrast]{\includegraphics[scale=0.5,width=0.16\textwidth]{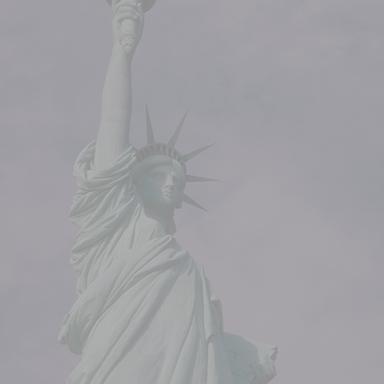}}\hskip.2em
    \subfloat[Fnoise]{\includegraphics[scale=0.5,width=0.16\textwidth]{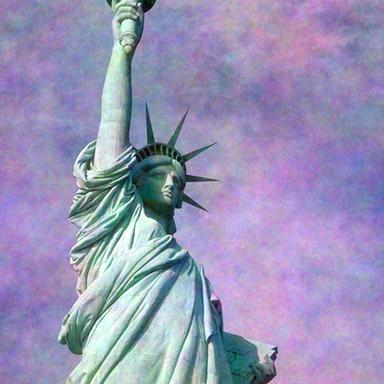}}\hskip.2em
    \subfloat[JPEG]{\includegraphics[scale=0.5,width=0.16\textwidth]{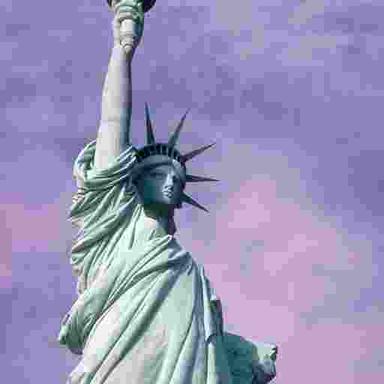}}\hskip.2em
    \subfloat[JPEG 2000]{\includegraphics[scale=0.5,width=0.16\textwidth]{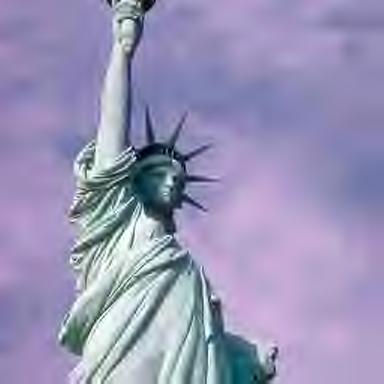}}
    \addtocounter{subfigure}{0}
    \subfloat[Reference image]{\includegraphics[scale=0.5,width=0.195\textwidth]{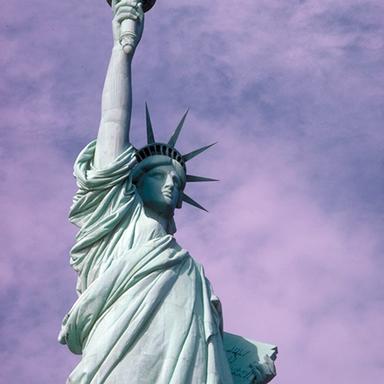}}\hskip.2em
    \subfloat[Translation (5\% pixels)]{\includegraphics[scale=0.5,width=0.195\textwidth]{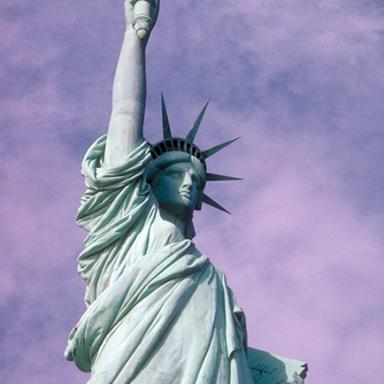}}\hskip.2em
    \subfloat[Rotation (3\degree) ]{\includegraphics[scale=0.5,width=0.195\textwidth]{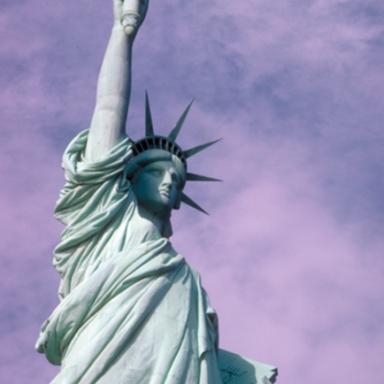}}\hskip.2em
    \subfloat[Scaling (1.05 factor)]{\includegraphics[scale=0.5,width=0.195\textwidth]{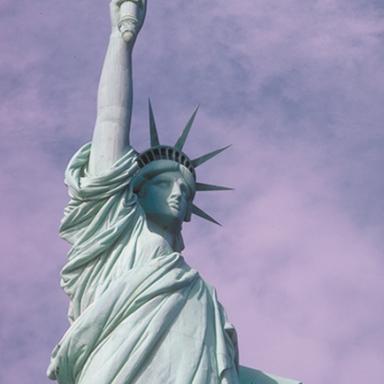}}\hskip.2em
    \subfloat[Combined]{\includegraphics[scale=0.5,width=0.195\textwidth]{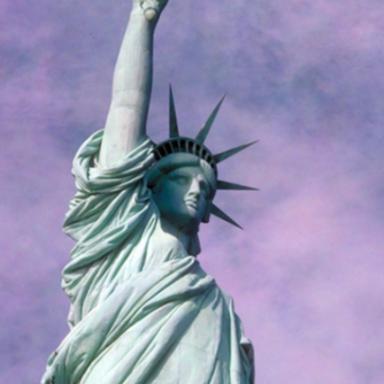}}
    \caption{Illustrations of the ``lady\_liberty" images in the CSIQ~\cite{larson2010most}. (a)~$\sim$~(f): Six distortion types used in the CSIQ~\cite{larson2010most}; (g) Reference image; (h)~$\sim$~(k): Four geometric transformations applied to create the CSIQ-GT dataset.}\label{fig:csiq_aug}
\end{figure*}

% Table generated by Excel2LaTeX from sheet 'table 4'
\begin{table*}[t]
  \centering
  \small
    \caption{Performance comparison of DeepDC against the state-of-the-art FR-IQA models on CSIQ-GT dataset. The best two results are highlighted in boldface and underline, respectively.}
  \label{tab:geo}%
    % Table generated by Excel2LaTeX from sheet 'table 4'
    \begin{tabular}{lcccccccccccccc}
    \toprule
    \multicolumn{1}{l}{\multirow{2}[4]{*}{Method}} & \multicolumn{2}{c}{Translation} && \multicolumn{2}{c}{Rotation} && \multicolumn{2}{c}{Scaling} && \multicolumn{2}{c}{Combined}   && \multicolumn{2}{c}{Overall} \\
    \cmidrule{2-3} \cmidrule{5-6} \cmidrule{8-9} \cmidrule{11-12}   \cmidrule{14-15}           & SRCC         & PLCC    &     & SRCC         & PLCC      &   & SRCC         & PLCC   &      & SRCC         & PLCC      &   & SRCC         & PLCC \\
    \midrule
    PSNR         & 0.166        & 0.267   &    & 0.177        & 0.248   &     & 0.179        & 0.259  &      & 0.158        & 0.196   &     & 0.070        & 0.392  \\
    SSIM~\cite{wang2004image} & 0.113        & 0.148   &     & 0.114        & 0.141  &      & 0.112        & 0.148    &    & 0.144        & 0.171    &    & 0.036        & 0.371  \\
    MS-SSIM~\cite{wang2003multiscale} & 0.119        & 0.150   &     & 0.123        & 0.150     &   & 0.120        & 0.222   &     & 0.148        & 0.165   &     & 0.038        & 0.389  \\
    VIF~\cite{sheikh2006image} & 0.159        & 0.315   &     & 0.163        & 0.273   &     & 0.158        & 0.296  &      & 0.183        & 0.307     &   & 0.137        & 0.302  \\
    FSIM~\cite{zhang2011fsim} & 0.135        & 0.222    &    & 0.128        & 0.196    &    & 0.130        & 0.190   &     & 0.110        & 0.149   &     & 0.076        & 0.359  \\
    NLPD~\cite{laparra2016perceptual} & 0.201        & 0.274   &     & 0.205        & 0.236   &     & 0.209        & 0.249   &     & 0.216        & 0.242    &    & 0.094        & 0.436  \\
    \midrule
    PieAPP~\cite{prashnani2018pieapp} & 0.378        & 0.576    &    & 0.462        & 0.616    &    & 0.359        & 0.574   &     & 0.192        & 0.502   &     & 0.412        & 0.540  \\
    LPIPS~\cite{zhang2018unreasonable} & 0.804        & 0.846    &    & 0.809        & 0.856   &     & 0.799        & 0.841   &     & 0.711        & 0.781   &     & 0.582        & 0.670  \\
    DSD~\cite{kligvasser2021deep} & 0.707        & 0.765   &     & 0.679        & 0.714     &   & 0.682        & 0.726  &      & 0.598        & 0.634 &       & 0.638        & 0.695  \\
    DISTS~\cite{ding2020image} & \underline{0.908}        & \underline{0.926}    &    & \underline{0.901}        & \underline{0.921}   &     & \underline{0.889}        & \underline{0.917}  &      & \underline{0.853}        & \underline{0.892}  &      & \underline{0.804}        & \underline{0.858}  \\
    DeepWSD~\cite{liao2022deepwsd} & 0.850        & 0.853    &    & 0.769        & 0.785    &    & 0.826        & 0.832   &     & 0.542        & 0.590   &     & 0.725        & 0.740  \\
    ST-LPIPS~\cite{ghildyal2022shift} & 0.533        & 0.549  &      & 0.462        & 0.492  &      & 0.502        & 0.529  &      & 0.223        & 0.379    &    & 0.376        & 0.433  \\
    \midrule
    DeepDC~(\textit{Ours}) & \textbf{0.929 } & \textbf{0.947 } && \textbf{0.922 } & \textbf{0.945 } && \textbf{0.915 } & \textbf{0.949 } && \textbf{0.893 } & \textbf{0.928 }& & \textbf{0.839 } & \textbf{0.891 } \\
    \bottomrule
    \end{tabular}%
    
\end{table*}%

\subsection{Performance on Texture Similarity Qualification}
\label{sec:texture}
To verify the effectiveness of our method on texture similarity qualification,
we conduct  experiments on two popular texture similarity datasets, \ie, SynTEX~\cite{golestaneh2015effect} and TQD~\cite{ding2020image}. Specifically, SynTEX consists of $105$ synthesized texture images, which were generated by five texture synthesis methods for $21$ high-quality texture images. TQD contains ten reference texture images, and each of them is degraded by 15 distortion types, including seven synthetic distortions, four texture synthesis methods~\cite{portilla2000parametric,gatys2015texture,snelgrove2017high,snelgrove2017high}, and four randomly resampling versions. For performance comparison, four representative knowledge-driven methods: SSIM~\cite{wang2004image}, VIF~\cite{sheikh2006image}, FSIM~\cite{zhang2011fsim}, and NLPD~\cite{laparra2016perceptual} and four data-driven methods: LPIPS~\cite{zhang2018unreasonable}, PieAPP~\cite{prashnani2018pieapp}, DISTS~\cite{ding2020image}, and ST-LPIPS~\cite{ghildyal2022shift} are selected. Furthermore, three models: STSIM~\cite{zujovic2013structural}, NPTSM~\cite{alfarraj2016content}, and IGSTOA~\cite{golestaneh2018synthesized}, that are designed especially for texture similarity are also included.

The SRCC and PLCC results are summarized in Table~\ref{tab:texutre_similarity}, from which we 
observe that the knowledge-driven FR-IQA models demonstrate inferior performance due to their reliance on pixel-by-pixel alignment. Besides, while the ST-LPIPS model is designed to accommodate mild pixel shifts, it fails to handle texture images effectively and performs even worse than the LPIPS model. In addition, ISGTQA exhibits noteworthy improvements for the texture similarity models but still falls behind DISTS and the proposed model. Overall, the proposed model produces consistent texture perceptual results in line with the HVS since it is able to measure both linear and nonlinear relationships, simultaneously.  
Thus, we may draw the conclusion that the proposed model fully utilizes the \textit{texture-sensitive} characteristic of pre-trained DNN features, providing a promising texture perception in the scenarios of texture similarity qualification.

\begin{figure*}
    \centering
    \captionsetup{justification=centering}
    \includegraphics[trim={1.2cm 2.5cm 0.5cm 2.5cm},clip,scale=0.5,width=0.98\textwidth]{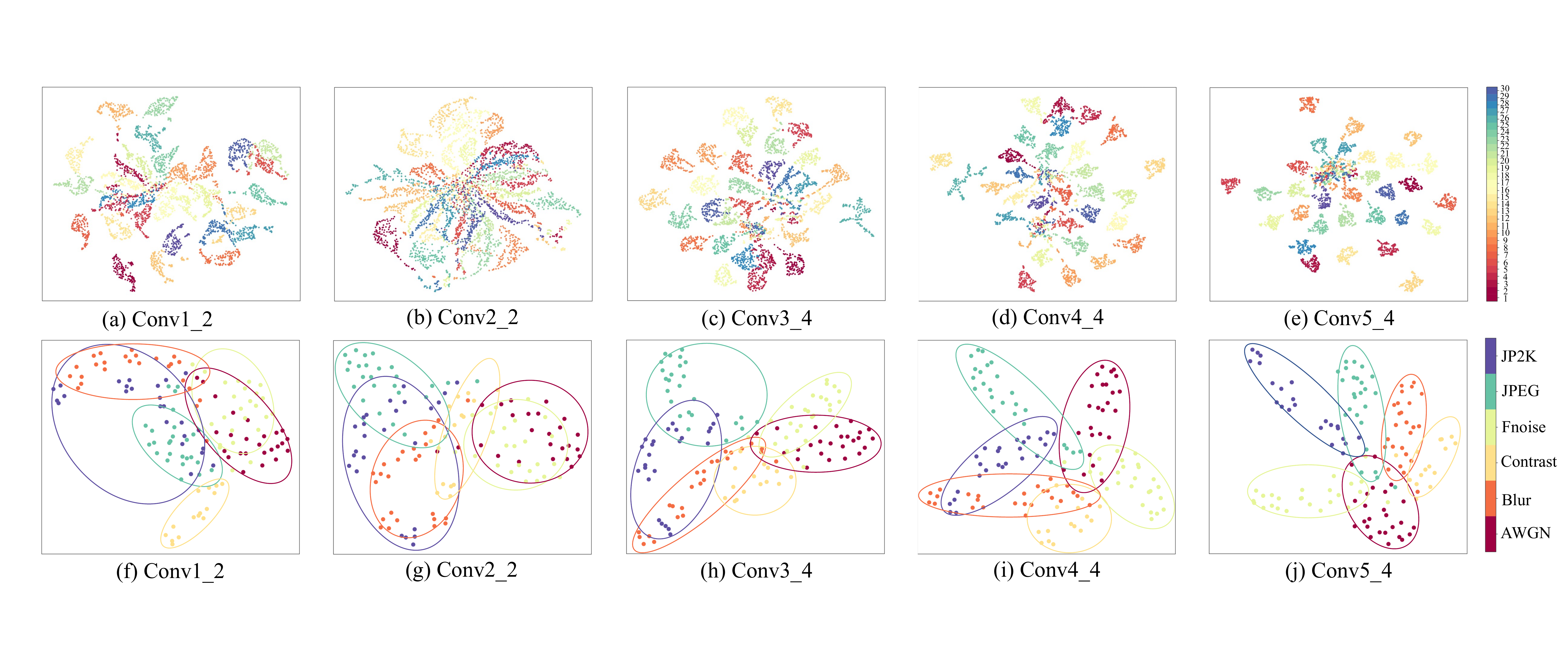}
    \caption{The UMAP visualization of the double-centered distance matrices based on five used convolutional layers. (a)~$\sim$~(e): The illustrations of $30$ image contents from shallow to deep layers on CSIQ-GT dataset; (f)~$\sim$~(j): The illustrations of six distortion types from shallow to deep layers of the image with the content of ``lady\_liberty" in CSIQ-GT dataset.}\label{fig:feature}
\end{figure*}

\subsection{Performance on Mild Geometric Transformation Assessment}

Texture image quality usually presents an invariance to mild geometric transformations~\cite{ding2020image}. To study the performance of existing FR-IQA models on such a prior, we construct a new IQA dataset based on the CSIQ dataset~\cite{larson2010most}, namely CSIQ geometric transformations (CSIQ-GT). The CSIQ contains $30$ reference images, which are degraded by six distortion types with three or four distortion levels. The six representative distortions are shown in Fig.~\ref{fig:csiq_aug}~(a)~$\sim$~(f), including additive white Gaussian noise~(AWGN), Gaussian blur~(Gblur), contrast, flicker noise~(Fnoise), JPEG, and JPEG~2000. We augment each distorted image with four geometric transformations, \ie, translation, rotation, scaling, and combined. In particular, the four geometric transformations are implemented by randomly shifting $5\%$ pixels in vertical or horizontal directions, randomly rotating $3^{\circ}$ in clockwise or anticlockwise directions, scaling the image by a factor of $1.05$, and mixing the above-mentioned transformations. One of the visual samples is shown in Fig.~\ref{fig:csiq_aug}~(h)~$\sim$~(k). We can observe the geometric transformations are almost imperceptible to HVS. We retain the source distorted images in the CSIQ-GT so that a total of $866 \times (4 + 1) = 4,330$ images are finally generated. As such, we make a mild assumption that the modest geometric transformations will not change the MOS of each image~\cite{ding2020image}. 
% The sample images are presented Fig.~\ref{fig:csiq_aug}. Then, we make a fair comparison on the CSIQ-GT dataset. 

We compare the proposed DeepDC to the same FR-IQA models as the standard quality evaluation on the constructed CSIQ-GT dataset. We list the SRCC and PLCC results for both overall and individual geometric transformation types in Table~\ref{tab:geo}. Our findings reveal a significant decrease in the performance of knowledge-driven models on the CSIQ-GT dataset. In contrast, our DeepDC model achieved superior performance across all four transformations. We believe this improvement lies in the globally linear and nonlinear statistic measure, which is capable of avoiding deterministic measurement of feature differences, exhibiting an invariance property. While ST-LPIPS is immune to $1$ or $2$ pixel shifts, it is vulnerable to more complex, unseen transformations. Moreover, through fine-tuning with large-scale texture images, DISTS is able to exhibit competitive performance in the presence of geometric transformations.

\subsection{Feature Visualization}
\label{sec:vis}
In order to elucidate the effectiveness of the proposed DeepDC quality model, we utilize the uniform manifold approximation and projection (UMAP)~\cite{mcinnes2018umap} to visualize the double-centered distance matrices for the distorted images in the constructed CSIQ-GT dataset. The projected features from five convolutional layers were analyzed regarding the image content and presented in Fig.~\ref{fig:feature}~(a)~$\sim$~(e), which yields several interesting observations. The double-centered distance matrices contain significant semantic information, and the features inherited from the deeper layers exhibit a well-defined clustering pattern based on different image contents. Then, our findings validate the efficacy of the proposed texture feature statistic-based model by utilizing the \textit{texture-sensitive} property~\cite{geirhos2018imagenet} of the pre-trained DNNs. Furthermore, we demonstrate the mapped two-dimensional embeddings based on the six distortion types for the image with the same content as Fig.~\ref{fig:csiq_aug}~(g)~(\ie, ``lady\_leberty"), which consists of a total of $140$ distorted images generated by six distortions and four geometric transformations. As shown in Fig.~\ref{fig:feature}~(f)~$\sim$~(j), the feature representation successfully clusters \bl{distorted} images according to their respective distortion types, even when geometric transformations are introduced. As such, the DeepDC enables the features to be equipped with meaningful semantic and quality information, the proposed model is capable of accurately predicting perceptual quality scores for imperceptible perturbation scenarios.

% Table generated by Excel2LaTeX from sheet 'ablations'
\begin{table*}[t]
  \centering
  \small 
      \caption{Ablation experiments of DeepDC with different DNN backbones and tailored at different layers in terms of SRCC results. The default backbone of DeepDC is represented in italics.}
    \begin{tabular}{lccccccccccc}
    \toprule
    \multirow{2}[4]{*}{Backbone} & \multirow{2}[4]{*}{Parameter} & \multicolumn{5}{c}{Quality prediction}                                   &              & \multicolumn{2}{c}{Texture similarity} &              & Geometric trans. \\
    \cmidrule{3-7}\cmidrule{9-10}\cmidrule{12-12}             &              & LIVE         & CSIQ         & TID2013      & KADID-10k    & PIPAL        &              & SynTEX       & TQD          &              & CSIQ-GT \\
    \midrule
    \multirow{2}[1]{*}{\textit{VGG19}} & Random       & 0.872        & 0.769        & 0.573        & 0.436        & 0.534        &              & 0.468        & 0.360        &              & 0.336  \\
                 & ImageNet     & \textbf{0.951 } & \textbf{0.951 } & 0.844        & \textbf{0.906 } & \textbf{0.684 } &              & 0.901        & \underline{0.889}         &              & 0.839  \\
                 \midrule
    \multirow{2}[0]{*}{ResNet50} & Random       & 0.860        & 0.787        & 0.589        & 0.490        & 0.580        &              & 0.485        & 0.390        &              & 0.378  \\
                 & ImageNet     & \underline{0.950}        & \underline{0.941}        & \underline{0.856}        & \underline{0.900}        & 0.674        &              & \textbf{0.933 } & 0.885        &              & \textbf{0.861 } \\
                 \midrule
    \multirow{2}[1]{*}{DenseNet121} & Random       & 0.854        & 0.757        & 0.601        & 0.576        & 0.572        &              & 0.461        & 0.631        &              & 0.405  \\
                 & ImageNet     & 0.948        & 0.934        & \textbf{0.857 } & 0.891        & \underline{0.677}        &              & \underline{0.922 }       & \textbf{0.918 } &              & \underline{0.860}  \\
    \bottomrule
    \end{tabular}%
        \begin{tablenotes}
      \scriptsize
      \item Geometric trans.: Geometric transformation.
    \end{tablenotes}
  \label{tab:ablation}%
\end{table*}%

\subsection{Ablation Studies}
\label{sec:ablation}
In this subsection, we conduct ablation experiments to investigate the impact of using different DNN backbones in the proposed model. Except for the default VGG19~\cite{simonyan2014very} backbone, we apply the proposed distance correlation measure to two other widely used image classification networks - ResNet50~\cite{he2016deep} and DenseNet121~\cite{huang2017densely}. Similarly, we calculate the Eqn.~(\ref{eq:quality}) from the coarse to fine convolutional layer for ResNet50 and DenseNet121.
In addition, we explore the effectiveness of pre-trained parameters by comparing results obtained with pre-trained and randomly initialized parameters. Our evaluation covers three quality assessment tasks, including image quality prediction, texture similarity qualification, and geometric transformation assessment. The results are summarized in Table~\ref{tab:ablation}, from which we find that the proposed DeepDC quality model is robust to changes in DNN backbones owing to its superior performance across all tested datasets. Besides, the backbones without pre-training~(\ie, random initialization) exhibit inferior performance on all datasets when computing the feature distance correlation. This outcome is expected, as these backbones do not contain accurate semantic and quality information. While both ResNet50 and DenseNet121 outperform VGG19 in some datasets~(TID2013, SynTEX, TQA, and CSIQ-GT), we ultimately select VGG19 as the default backbone due to its satisfactory trade-off between model complexity and performance.

\section{Applications}\label{sec:app}
% In this section, we take the proposed DeepDC as an objective function to globally optimized for texture synthesis
In this section, we further demonstrate the effectiveness of the proposed FR-IQA model with the analysis-by-synthesis evaluations~\cite{julesz1962visual,portilla2000parametric}. we first optimize the proposed DeepDC model for texture synthesis in the image space. Subsequently, we take the proposed DeepDC as an objective function for the task of real-time neural style transfer~(NST).

\subsection{Texture Synthesis}

\begin{figure*}
    \centering
    \captionsetup{justification=centering}

    \subfloat{\includegraphics[scale=0.5,width=0.16\textwidth]{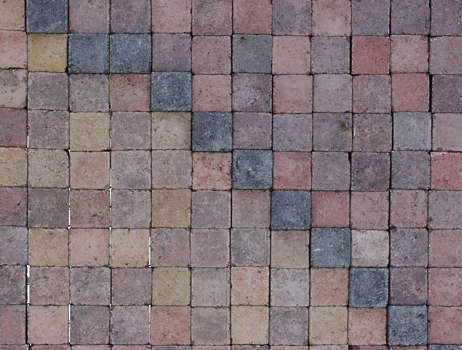}}\hskip.2em
    \subfloat{\includegraphics[scale=0.5,width=0.16\textwidth]{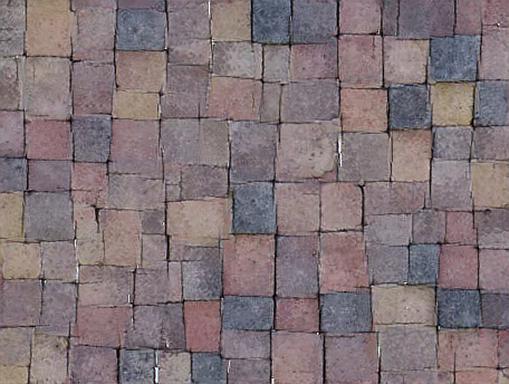}}\hskip.2em    
    \subfloat{\includegraphics[scale=0.5,width=0.16\textwidth]{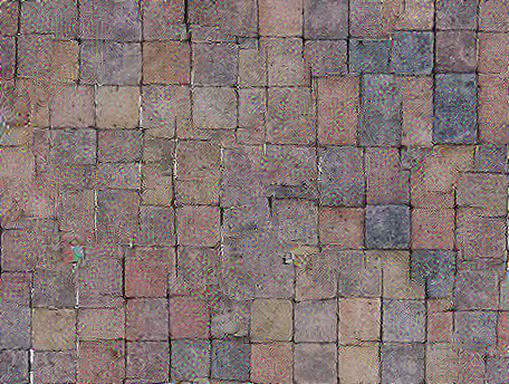}}\hskip.2em    
    \subfloat{\includegraphics[scale=0.5,width=0.16\textwidth]{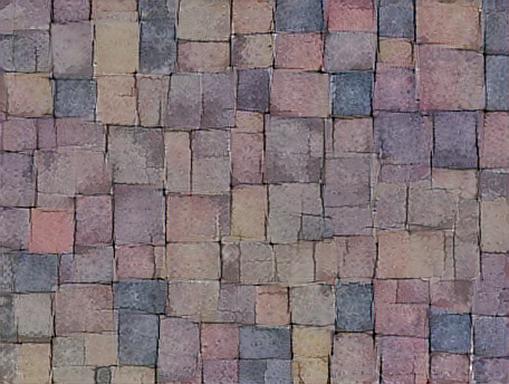}}\hskip.2em    
    \subfloat{\includegraphics[scale=0.5,width=0.16\textwidth]{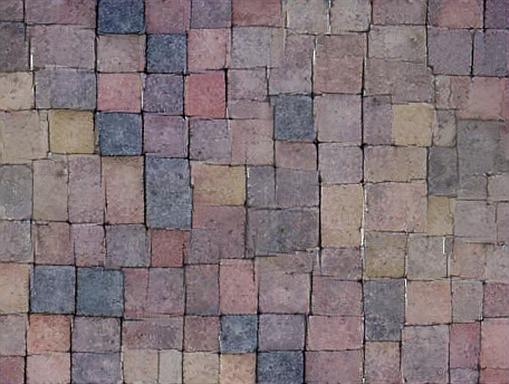}}\hskip.2em    
    \subfloat{\includegraphics[scale=0.5,width=0.16\textwidth]{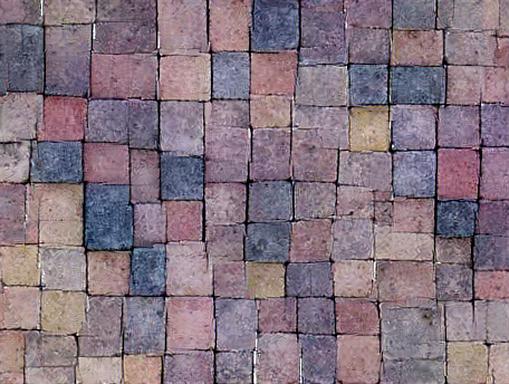}}\\
    \addtocounter{subfigure}{-6}
    \vspace{-5pt}
    \subfloat{\includegraphics[scale=0.5,width=0.16\textwidth]{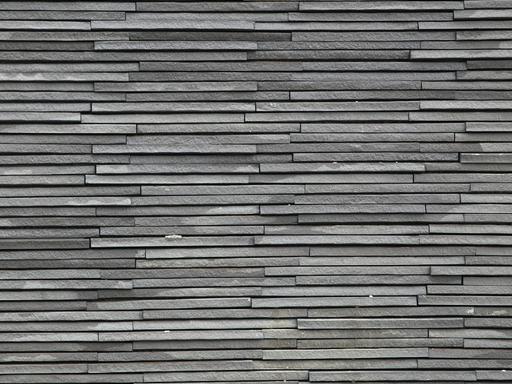}}\hskip.2em
    \subfloat{\includegraphics[scale=0.5,width=0.16\textwidth]{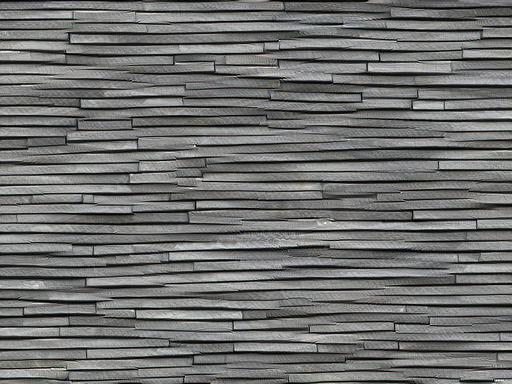}}\hskip.2em    
    \subfloat{\includegraphics[scale=0.5,width=0.16\textwidth]{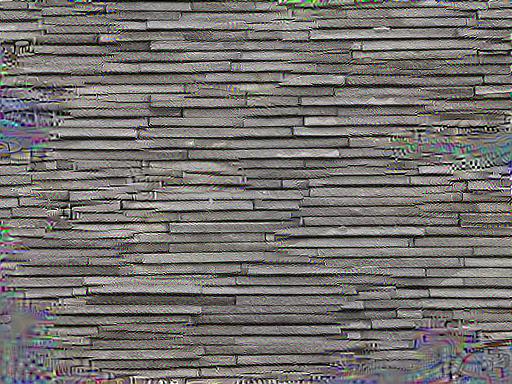}}\hskip.2em    
    \subfloat{\includegraphics[scale=0.5,width=0.16\textwidth]{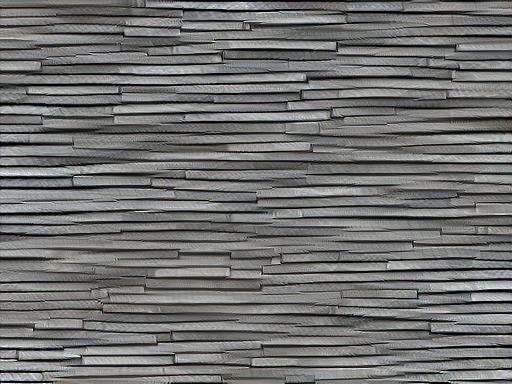}}\hskip.2em    
    \subfloat{\includegraphics[scale=0.5,width=0.16\textwidth]{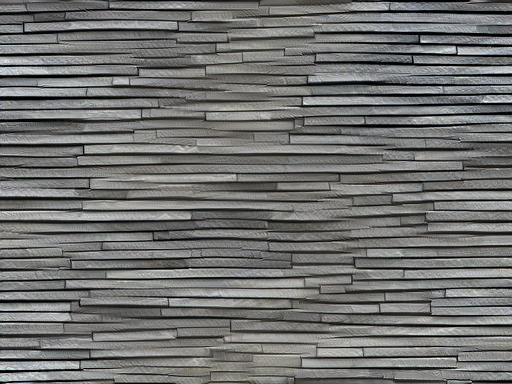}}\hskip.2em    
    \subfloat{\includegraphics[scale=0.5,width=0.16\textwidth]{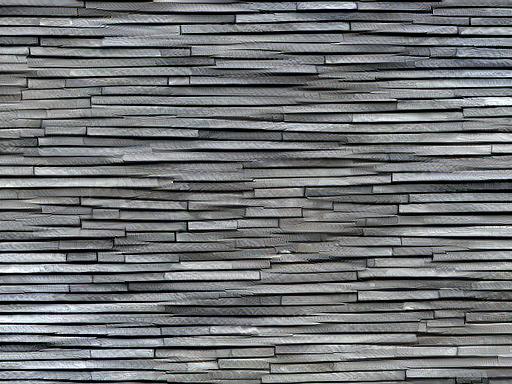}}\\
    \addtocounter{subfigure}{-6}
    \vspace{-5pt}
    \subfloat{\includegraphics[scale=0.5,width=0.16\textwidth]{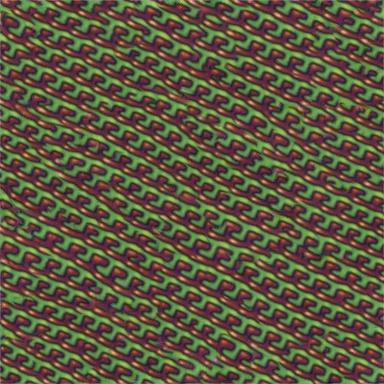}}\hskip.2em
    \subfloat{\includegraphics[scale=0.5,width=0.16\textwidth]{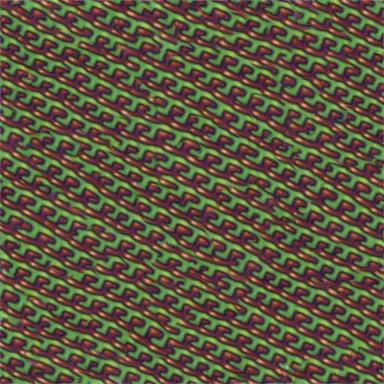}}\hskip.2em    
    \subfloat{\includegraphics[scale=0.5,width=0.16\textwidth]{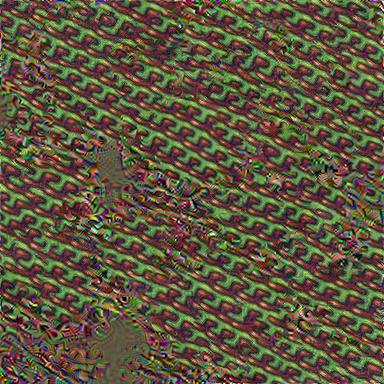}}\hskip.2em    
    \subfloat{\includegraphics[scale=0.5,width=0.16\textwidth]{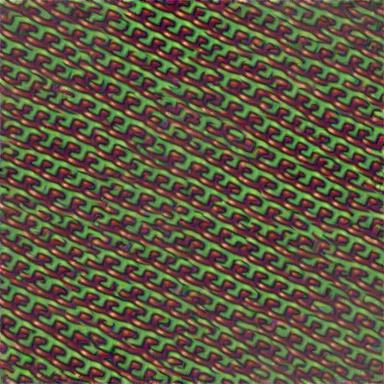}}\hskip.2em    
    \subfloat{\includegraphics[scale=0.5,width=0.16\textwidth]{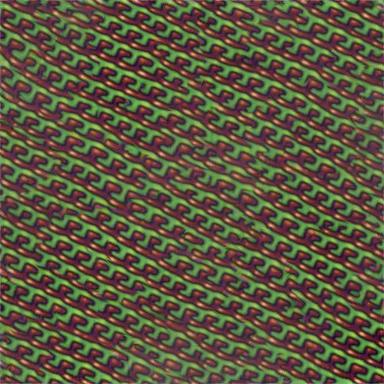}}\hskip.2em    
    \subfloat{\includegraphics[scale=0.5,width=0.16\textwidth]{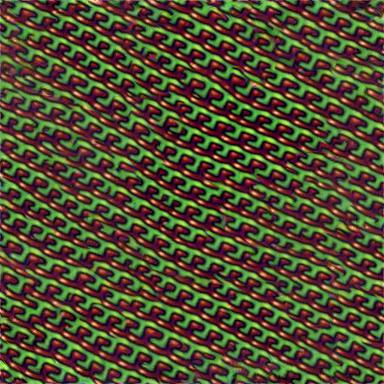}}\\
    \addtocounter{subfigure}{-6}
    \vspace{-5pt}
    \subfloat[Source image]{\includegraphics[scale=0.5,width=0.16\textwidth]{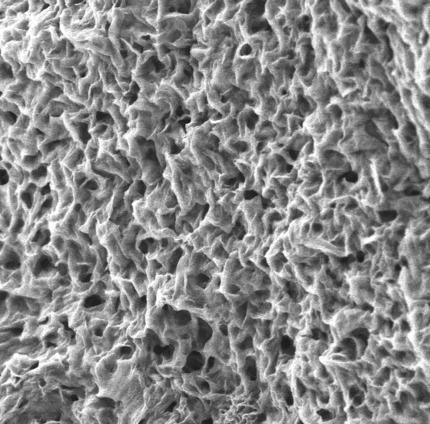}}\hskip.2em
    \subfloat[Gatys~\etal~\cite{gatys2015texture}]{\includegraphics[scale=0.5,width=0.16\textwidth]{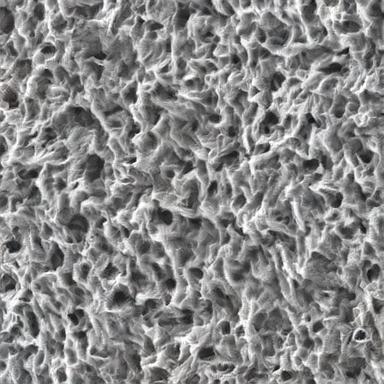}}\hskip.2em    
    \subfloat[CXT~\cite{mechrez2018contextual}]{\includegraphics[scale=0.5,width=0.16\textwidth]{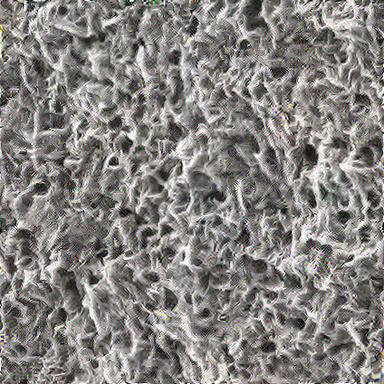}}\hskip.2em    
    \subfloat[DISTS~\cite{ding2020image}]{\includegraphics[scale=0.5,width=0.16\textwidth]{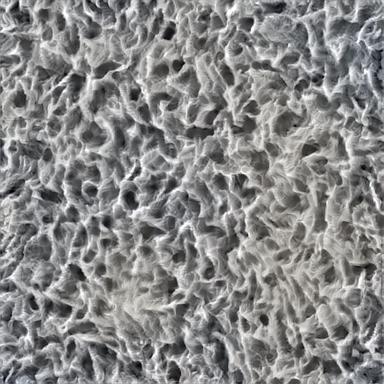}}\hskip.2em    
    \subfloat[SWD~\cite{heitz2021sliced}]{\includegraphics[scale=0.5,width=0.16\textwidth]{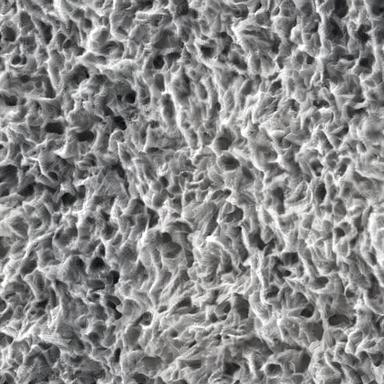}}\hskip.2em    
    \subfloat[DeepDC (\textit{Ours})]{\includegraphics[scale=0.5,width=0.16\textwidth]{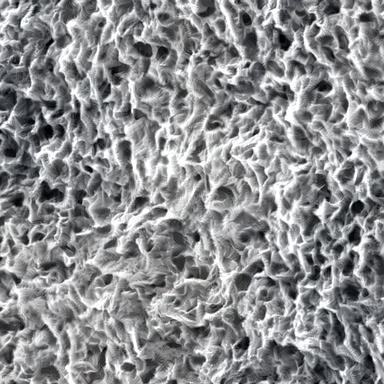}}\\
    \caption{Illustration of the qualitative comparisons for the DeepDC against Gatys~\etal~\cite{gatys2015texture}, CXT~\cite{mechrez2018contextual}, DISTS~\cite{ding2020image}, and SWD~\cite{heitz2021sliced}.}\label{fig:texture}
\end{figure*}

Texture synthesis involves the generation of an image that matches the statistical properties of a given texture photograph. The visual attributes of textures are commonly characterized by local statistical measures~\cite{portilla2000parametric,heitz2021sliced,mechrez2018contextual}. If the statistical measures accurately represent the visual appearance of the texture, the synthesized image should be perceptually identical to the reference photograph~\cite{julesz1962visual,guo2010completed,dong2020perceptual}. Texture synthesis can be roughly divided into parametric and non-parametric methods.
Parametric texture synthesis uses a compact parametric model, learned from example textures, to generate new texture samples with explicit control over properties like color, size, shape, and orientation~\cite{portilla2000parametric,wei2000fast,gatys2015texture,mechrez2018contextual,ding2020image,heitz2021sliced}. Non-parametric texture synthesis directly samples and copies pixel values from the example texture to reproduce complex textures with high fidelity, but lacks explicit control over texture properties~\cite{efros1999texture,kwatra2005texture}. 

In this experiment, we focus on the parametric methods based on the ImageNet pre-trained VGG network, which shares the same feature descriptor as the DeepDC model. However, the main difference between the competing and proposed methods lies in diverse distribution-based objectives. Overall, optimizing the following objective in the image space is used to synthesize texture images, 
\begin{align}\label{eq:opt}
    \bm y^* = \argmin_{\bm y}\mathcal{T}(\bm x, \bm y),
\end{align}
where $\bm x$ represents the reference texture image, $\bm y^*$ is the optimized texture image, which is generated by the gradient descent optimization from random initialization, and $\mathcal{T}$ is the various objective functions.

\noindent\textbf{Selected Methods.} We compare the proposed model to four representative statistics with the similar feature descriptor~\cite{simonyan2014very}, including,
\begin{itemize}
    \item Gatys~\etal~\cite{gatys2015texture}: Gatys~\etal employed the Gram matrix to establish a better texture synthesis method, which was implemented by computing the inner product among several layers of the VGG feature maps. \item CXT~\cite{mechrez2018contextual}: The contextual loss was further proposed to measure the statistical similarity between the reference and optimized texture images. 
    \item DISTS~\cite{ding2020image}: Ding~\etal found that taking the average of the pre-trained feature representations is adequate to model a variety of visual textures. 
    \item SWD~\cite{heitz2021sliced}: Heitz~\etal optimized the Sliced Wasserstein distance of the deep VGG features to capture full feature distributions, resulting in better texture perception quality.
\end{itemize}

\noindent\textbf{Dataset.}\label{sec:texture_synthesis}
 We construct a texture dataset from Pixelbay\footnote{\url{https://pixabay.com/images/search/texture/}}, which contains $30$ classical texture images. The images cover a variety of texture patterns, such as wood, grain, stone and rock, fabric and textiles,  metal, leather, paper, and water. Several sampled images are shown in Fig.~\ref{fig:texture}~(a).

\begin{figure}
    \centering
    \captionsetup{justification=centering}
    \includegraphics[trim={0cm 0cm 0cm 0cm},clip,scale=0.5,width=0.48\textwidth]{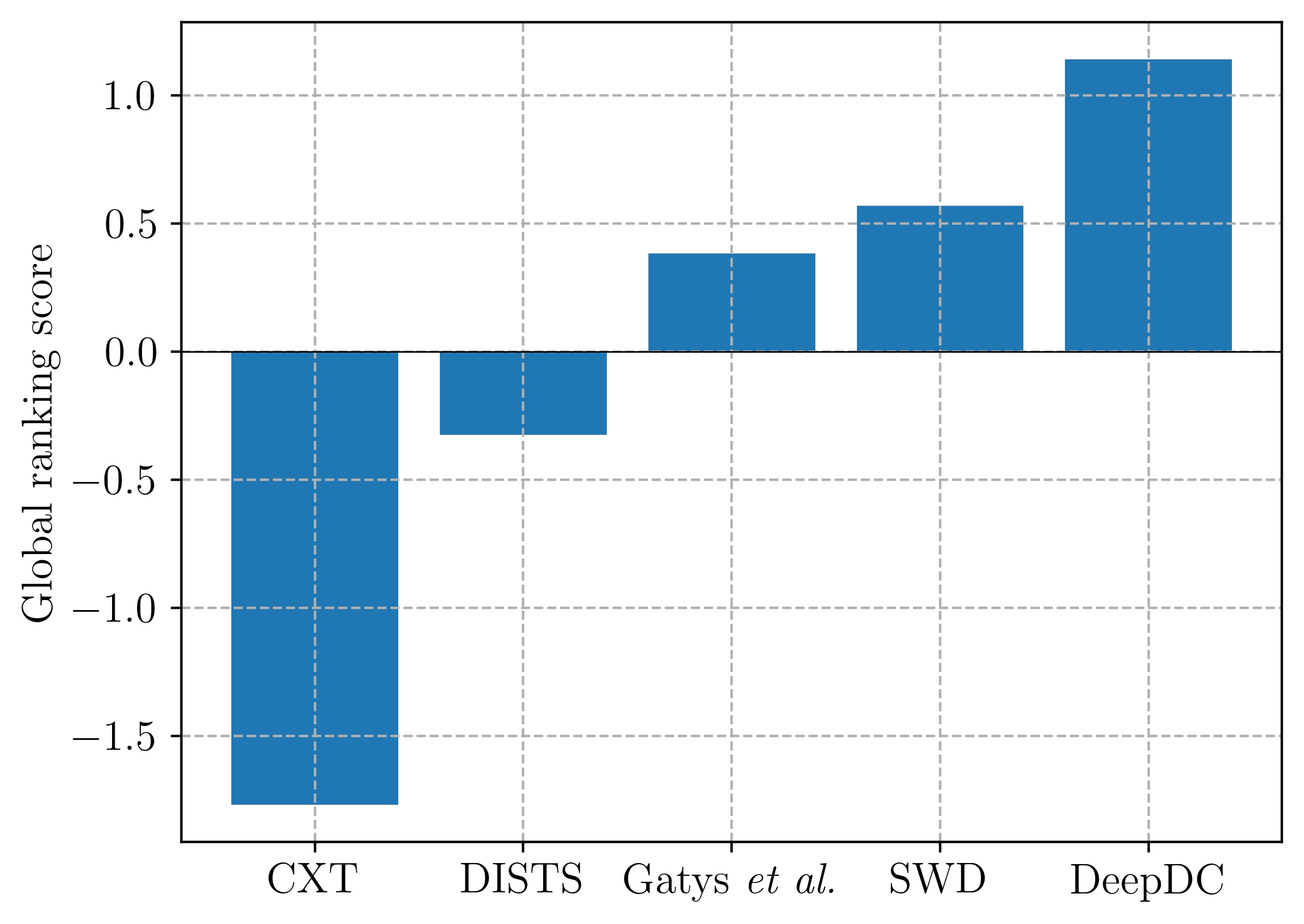}
    \caption{Illustration of the global ranking scores between the proposed method and four texture synthesis methods using the  Bradley-Terry model~\cite{Tsukida2011}.}\label{fig:texture_score}
\end{figure}
 
\noindent\textbf{Experimental Settings.} We apply their original implementations to reproduce the result of the four competing methods. We perform the texture optimization with an L-BFGS-B optimizer~\cite{byrd1995limited}, and the learning rate is fixed to $1$. The final output of each algorithm is determined by its convergence after 50 optimization iterations. As such, a total of $150$ texture images are generated using the proposed and four state-of-the-art texture synthesis methods. To make a fair comparison for the optimized texture images, we carry out subjective user testing using the two-alternative forced choice (2AFC) method. We invite $20$ subjects to perform the subjective experiment in a controllable indoor environment with normal light, utilizing a calibrated true-color LCD monitor in compliance with ITU-T BT.500 recommendations~\cite{video2000final}.

\begin{figure*}
    \centering
    \captionsetup{justification=centering}
    \subfloat{\includegraphics[scale=0.5,width=0.14\textwidth]{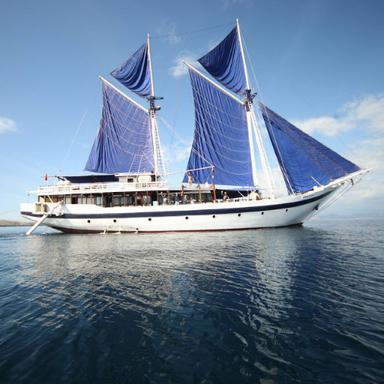}}\hskip.1em
    \subfloat{\includegraphics[scale=0.5,width=0.14\textwidth]{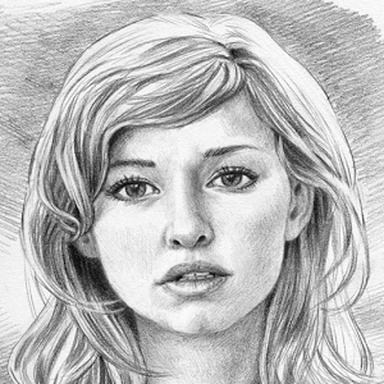}}\hskip.1em
    \subfloat{\includegraphics[scale=0.5,width=0.14\textwidth]{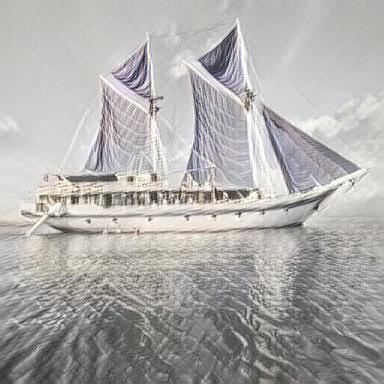}}\hskip.1em
    \subfloat{\includegraphics[scale=0.5,width=0.14\textwidth]{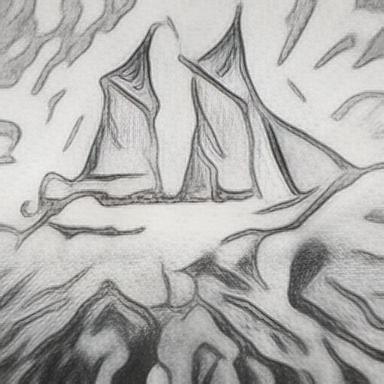}}\hskip.1em
    \subfloat{\includegraphics[scale=0.5,width=0.14\textwidth]{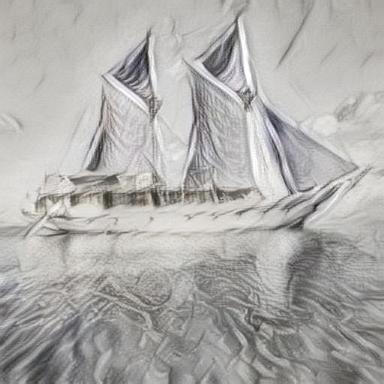}}\hskip.1em
    \subfloat{\includegraphics[scale=0.5,width=0.14\textwidth]{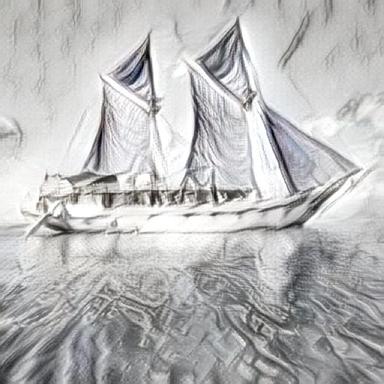}}\hskip.1em
    \subfloat{\includegraphics[scale=0.5,width=0.14\textwidth]{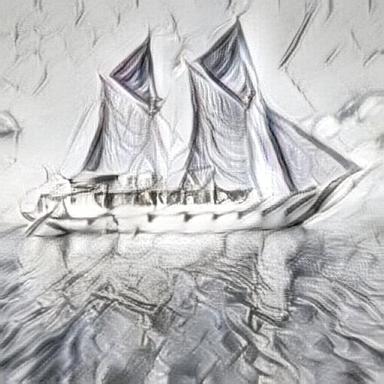}}\hskip.1em\\
    \addtocounter{subfigure}{-7}
    \vspace{-5pt}
    \subfloat{\includegraphics[scale=0.5,width=0.14\textwidth]{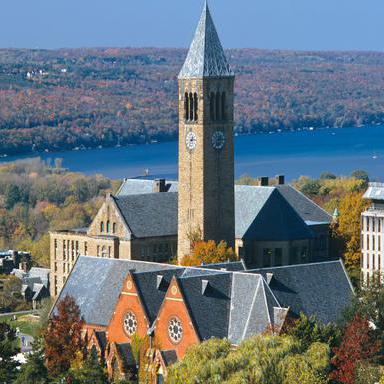}}\hskip.1em
    \subfloat{\includegraphics[scale=0.5,width=0.14\textwidth]{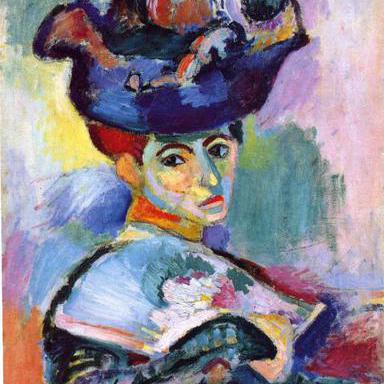}}\hskip.1em
    \subfloat{\includegraphics[scale=0.5,width=0.14\textwidth]{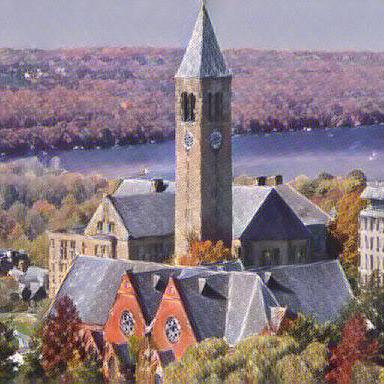}}\hskip.1em
    \subfloat{\includegraphics[scale=0.5,width=0.14\textwidth]{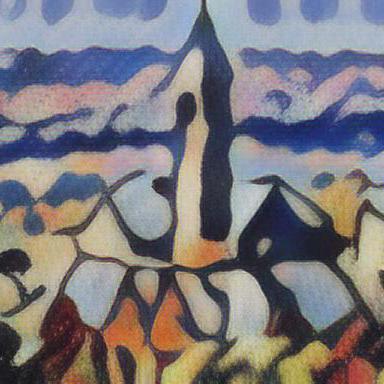}}\hskip.1em
    \subfloat{\includegraphics[scale=0.5,width=0.14\textwidth]{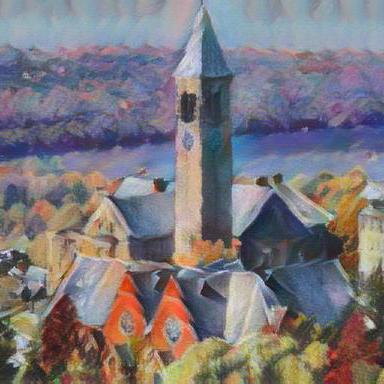}}\hskip.1em
    \subfloat{\includegraphics[scale=0.5,width=0.14\textwidth]{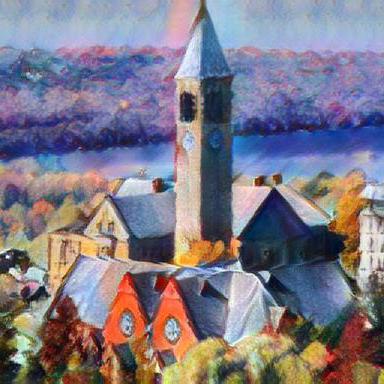}}\hskip.1em
    \subfloat{\includegraphics[scale=0.5,width=0.14\textwidth]{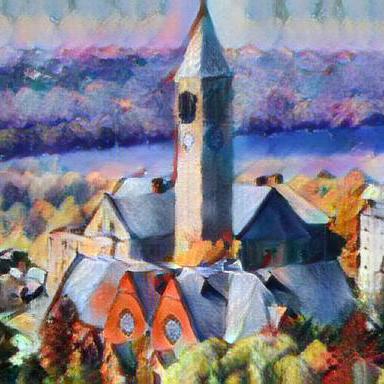}}\hskip.1em\\
    \addtocounter{subfigure}{-7}
    \vspace{-5pt}
    \subfloat{\includegraphics[scale=0.5,width=0.14\textwidth]{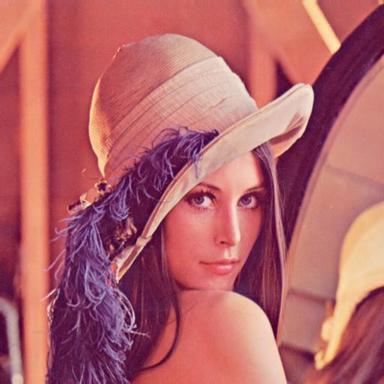}}\hskip.1em
    \subfloat{\includegraphics[scale=0.5,width=0.14\textwidth]{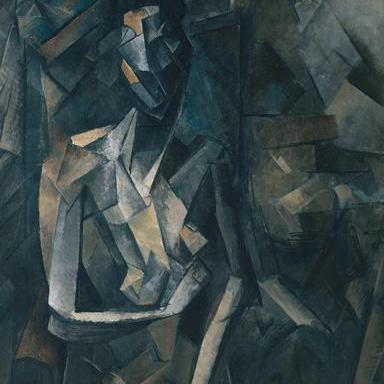}}\hskip.1em
    \subfloat{\includegraphics[scale=0.5,width=0.14\textwidth]{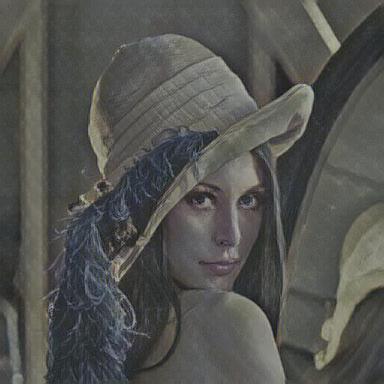}}\hskip.1em
    \subfloat{\includegraphics[scale=0.5,width=0.14\textwidth]{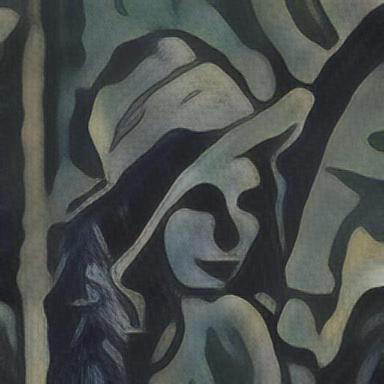}}\hskip.1em
    \subfloat{\includegraphics[scale=0.5,width=0.14\textwidth]{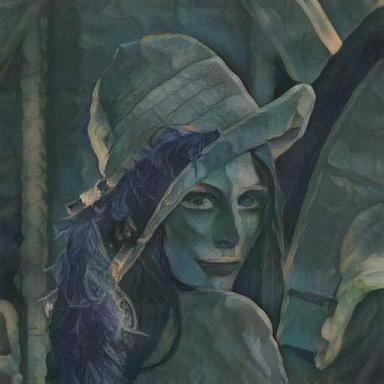}}\hskip.1em
    \subfloat{\includegraphics[scale=0.5,width=0.14\textwidth]{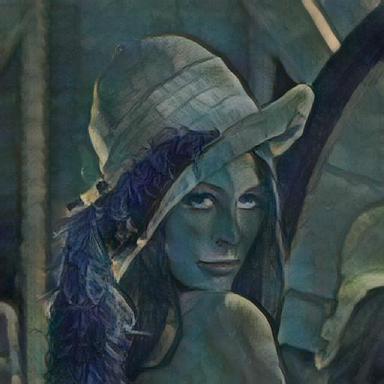}}\hskip.1em
    \subfloat{\includegraphics[scale=0.5,width=0.14\textwidth]{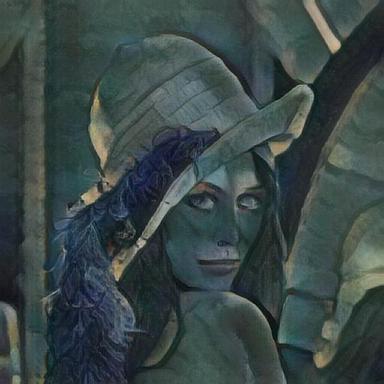}}\hskip.1em\\
    \addtocounter{subfigure}{-7}
    \vspace{-5pt}
    \subfloat{\includegraphics[scale=0.5,width=0.14\textwidth]{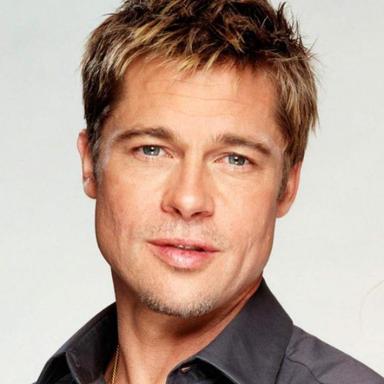}}\hskip.1em
    \subfloat{\includegraphics[scale=0.5,width=0.14\textwidth]{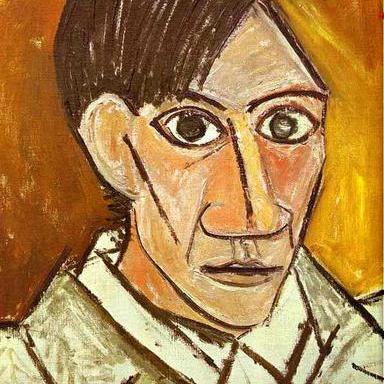}}\hskip.1em
    \subfloat{\includegraphics[scale=0.5,width=0.14\textwidth]{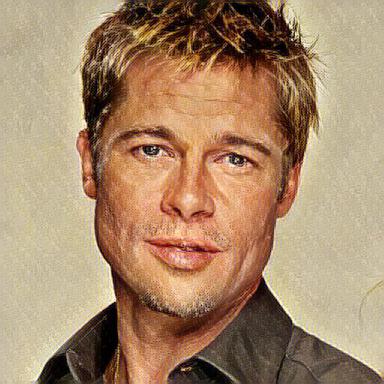}}\hskip.1em
    \subfloat{\includegraphics[scale=0.5,width=0.14\textwidth]{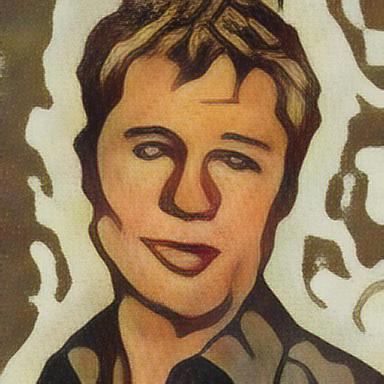}}\hskip.1em
    \subfloat{\includegraphics[scale=0.5,width=0.14\textwidth]{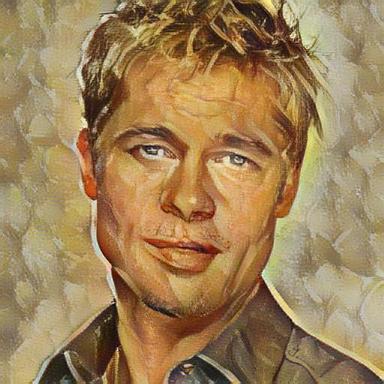}}\hskip.1em
    \subfloat{\includegraphics[scale=0.5,width=0.14\textwidth]{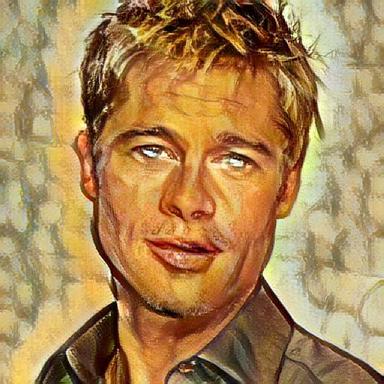}}\hskip.1em
    \subfloat{\includegraphics[scale=0.5,width=0.14\textwidth]{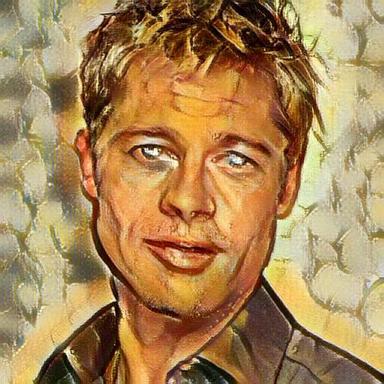}}\hskip.1em\\
    \addtocounter{subfigure}{-7}
    \vspace{-5pt}
    \subfloat[Content]{\includegraphics[scale=0.5,width=0.14\textwidth]{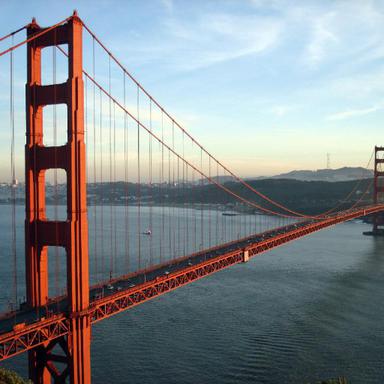}}\hskip.1em
    \subfloat[Style]{\includegraphics[scale=0.5,width=0.14\textwidth]{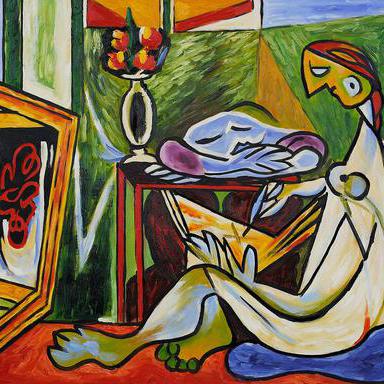}}\hskip.1em
    \subfloat[Johnson~\etal~\cite{johnson2016perceptual}]{\includegraphics[scale=0.5,width=0.14\textwidth]{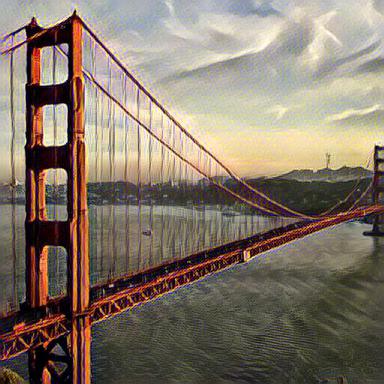}}\hskip.1em
    \subfloat[CXT~\cite{mechrez2018contextual}]{\includegraphics[scale=0.5,width=0.14\textwidth]{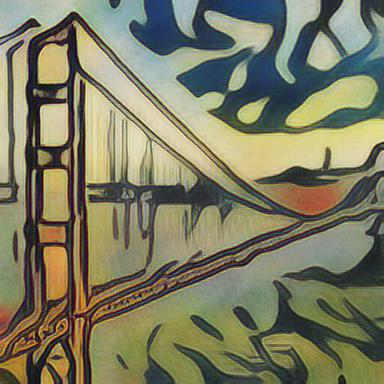}}\hskip.1em
    \subfloat[AdaIN~\cite{huang2017arbitrary}]{\includegraphics[scale=0.5,width=0.14\textwidth]{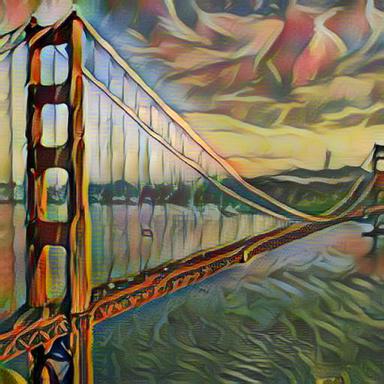}}\hskip.1em
    \subfloat[Li~\etal~\cite{li2022sliced}]{\includegraphics[scale=0.5,width=0.14\textwidth]{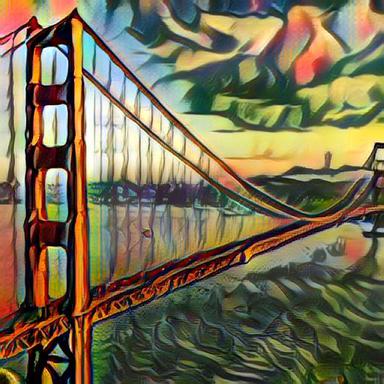}}\hskip.1em
    \subfloat[DeepDC~(\textit{Ours})]{\includegraphics[scale=0.5,width=0.14\textwidth]{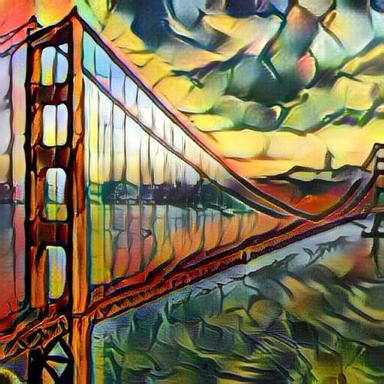}}\hskip.1em\\
    % \addtocounter{subfigure}{-7}
    % \vspace{-5pt}
    \caption{Illustration of the qualitative comparisons for the DeepDC against Johnson~\etal~\cite{johnson2016perceptual}, CXT~\cite{mechrez2018contextual}, AdaIN~\cite{huang2017arbitrary}, and Li~\etal~\cite{li2022sliced}.}\label{fig:style_transfer}
\end{figure*}

\noindent\textbf{Performance Comparisons.}
We present several optimized texture images in Fig.~\ref{fig:texture}. In particular, the images optimized by Gatys~\etal~\cite{heitz2021sliced} and DISTS~\cite{ding2020image} fail to generate structure details, resulting in a blurry appearance. CXT~\cite{mechrez2018contextual} suffers from apparent noise artifacts because of performing contextual similarity on high-level feature representations. Though SWD demonstrates competitive performance with satisfied results, it still underperforms the proposed method. In contrast, the proposed method generates visually appealing texture images with faithful details and vivid color. In addition, we leverage the Bradley-Terry model~\cite{Tsukida2011} to analyze paired comparison data from the subjective user study. The results are shown in Fig.~\ref{fig:texture_score}, from which our method outperforms the comparison methods with a significant margin, whose success may be attributed to the linear and nonlinear measure.  CXT~\cite{mechrez2018contextual} is subpar in our test because of the failure to suppress speckle noise.

\subsection{Neural Style Transfer}
Real-time neural style transfer (NST) is an image generation technique that employs DNNs to merge the content of one image with the style of another~\cite{jing2019neural}. NST enables flexible, automated, and highly effective style transfer between images by utilizing DNNs for style and content representations. The content and style are the two most crucial factors in determining the final results. For the content term, Johnson~\etal~\cite{johnson2016perceptual} proposed the widely adopted perceptual loss to preserve the objects, shapes, and structures in the image. Additionally, researchers have devoted considerable effort to capturing textures and color patterns from style images, including various statistical measures~\cite{johnson2016perceptual,li2017universal,mechrez2018contextual}, distribution matching methods~\cite{risser2017stable,zhang2022exact,li2022sliced}, and GAN-based methods~\cite{kotovenko2019content,azadi2018multi}.

In this work, we demonstrate that DeepDC is sufficient as a style objective for transferring texture and color information. Consequently, We integrate DeepDC into the adaptive instance normalization~(AdaIN)~\cite{huang2017arbitrary} framework for a novel NST method. Specifically, AdaIN performs style transfer by matching feature statistics between content and style images using adaptive instance normalization in an encoder-decoder architecture~\cite{huang2017arbitrary}. The overall loss function $\mathcal{L}$ can be formulated as follows,
\begin{align}\label{eq:loss}
    \mathcal{L} = \mathcal{L}_c + \lambda\mathcal{L}_s, 
\end{align}
where $\mathcal{L}_c$ and $ \mathcal{L}_s$ represent the content and style functions, and $\lambda$ is a balance weight for those two terms.
We replace the mean and standard deviation loss $ \mathcal{L}_s$ of AdaIN with the DeepDC~(Eqn.~(\ref{eq:quality})) without any other modifications.

\noindent\textbf{Selected Methods.} We compare our proposed method to four advanced NST approaches~\cite{johnson2016perceptual,mechrez2018contextual,huang2017arbitrary,li2022sliced}, which rely on different statistical measures to model style information. Specifically,

\begin{itemize}
    \item Johnson~\etal~\cite{johnson2016perceptual}: The perceptual losses and Gram matrix were first used to improve the quality in real-time style transfer and image super-resolution.
    \item CXT~\cite{mechrez2018contextual}: Chen~\etal proposed a contextual loss function to measure visual similarity for image transformation tasks using non-aligned training data.
    \item AdaIN~\cite{huang2017arbitrary}: Huang~\etal presented to use an adaptive instance normalization to match the mean and stand deviation feature statistics between content and style images for style transfer.
    \item Li~\etal~\cite{li2022sliced}: A novel sliced Wasserstein distance based approach was proposed to  enhance the quality of the style transfer images.
\end{itemize}

\begin{figure}
    \centering
    \captionsetup{justification=centering}
    \includegraphics[trim={0cm 0cm 0cm 0cm},clip,scale=0.5,width=0.48\textwidth]{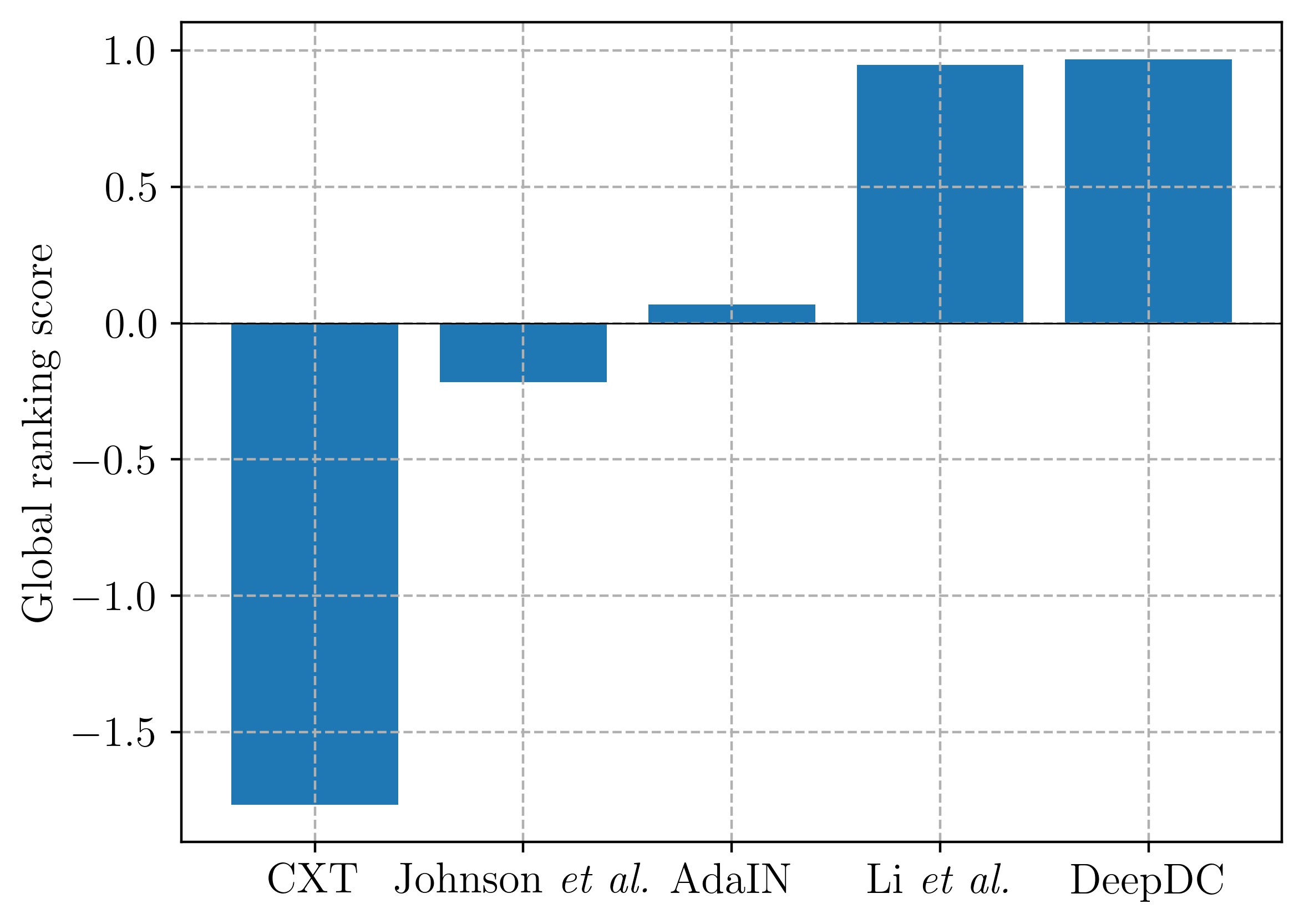}
    \caption{Illustration of the global ranking scores between the propose method and four NST algorithms using the  Bradley-Terry model~\cite{Tsukida2011}.}\label{fig:style_score}
\end{figure}

\noindent\textbf{Dataset.}
In order to train our model, we employ the training set of MS-COCO~\cite{lin2014microsoft} as the source of content and the WikiArt training set~\cite{wikiart2016} as the source of style. The MS-COCO dataset comprises approximately $80,000$ images depicting common objects in context, and the WikiArt training set contains around $80,000$ artistic images. During the training process, the images are initially resized to dimensions of $512\times512\times3$, and subsequently subjected to random cropping to produce $256\times256\times3$  patches. The test set utilized in AdaIN~\cite{huang2017arbitrary} is adopted as our test set, which consists of 11 content images and 20 style images, respectively. We apply the same training and testing sets to the comparison methods with their original implementations.

\noindent\textbf{Experimental Settings.} We use Adam~\cite{Kingma2014adam} with a learning rate $1\times10^{-4}$ to optimize the decoder using the overall loss function~(Eqn.~(\ref{eq:loss})). We set the batch size to $48$ and the training iteration to $160,000$. We empirically fix $\lambda=10$ to balance the two loss terms. The perceptual loss is computed on \texttt{conv4\_4} of a pre-trained VGG19
model and the style loss is conducted over \texttt{conv1\_2}, \texttt{conv2\_2}, \texttt{conv3\_4}, \texttt{conv4\_4}. In addition, we conduct a 2AFC subjective test to compare our proposed method with four competing methods. \bl{The subjective experiment is carried out in a controllable indoor environment with normal light, and the other $20$ subjects are invited to participate in the experiments~\cite{video2000final}.}

\noindent\textbf{Performance Comparisons.}
As illustrated in Fig.~\ref{fig:style_transfer}, a series of five content and style images are presented, complemented by the stylized images produced by our method and four competing methods. As shown in Fig.~\ref{fig:style_transfer} (c), Johnson~\etal~\cite{johnson2016perceptual} overemphasizes the preservation of content information, which subsequently introduces noticeable noise artifacts. In contrast, the CXT method \cite{mechrez2018contextual} lacks the ability to generate distinct structural details. Although AdaIN~\cite{huang2017arbitrary} and Li~\etal~\cite{li2022sliced} produce competitive results, AdaIN~\cite{huang2017arbitrary} exhibits a pale visual appearance, while Li~\etal~\cite{li2022sliced} suffers from halo artifacts around regions with strong edges.  On the other hand, the proposed method successfully avoids these artifacts and effectively transfers structural details and color patterns, resulting in the highest visual quality. Furthermore, a quantitative evaluation using the Bradley-Terry model on paired comparison data further indicates the proposed method achieves the highest average ranking score among the five style transfer methods, as shown in Fig.~\ref{fig:style_score}. Besides, another advanced texture statistic-based method~(\ie, Li~\etal~\cite{li2022sliced}) demonstrates the competitive performance with the proposed method, indicating the improvement of the representation capability with more robust texture descriptors.

\section{Conclusion}
In this paper, we have presented a novel FR-IQA model that fully leverages the \textit{texture-sensitive} characteristic of the pre-trained DNN backbones. Our model calculates distance correlation in the domain of pre-trained deep features, eliminating the need for MOS fine-tuning. \red{It is worth noting that the model is able to measure both linear and nonlinear relationships of the deep features. }
These advantageous properties equip DeepDC to function effectively as a perceptual quality evaluator, as evidenced by its superior performance across five standard IQA datasets, a perceptual similarity dataset, two texture similarity datasets, and a geometric transformations dataset. \red{Apart from} its effectiveness as an FR-IQA model, we demonstrated its utility in synthesizing texture images and serving as a style loss function for NST. Both quantitative and qualitative experiments validate the proposed model produces visually appealing results.  

\appendices

% % you can choose not to have a title for an appendix
% % if you want by leaving the argument blank
% \section{}

% \ifCLASSOPTIONcompsoc
%   % The Computer Society usually uses the plural form
%   \section*{Acknowledgments}
% \else
%   % regular IEEE prefers the singular form
% \fi


\ifCLASSOPTIONcaptionsoff
  \newpage
\fi

\bibliographystyle{IEEEtran}\
\small{
\bibliography{ref}}

\begin{thebibliography}{10}
\providecommand{\url}[1]{#1}
\csname url@samestyle\endcsname
\providecommand{\newblock}{\relax}
\providecommand{\bibinfo}[2]{#2}
\providecommand{\BIBentrySTDinterwordspacing}{\spaceskip=0pt\relax}
\providecommand{\BIBentryALTinterwordstretchfactor}{4}
\providecommand{\BIBentryALTinterwordspacing}{\spaceskip=\fontdimen2\font plus
\BIBentryALTinterwordstretchfactor\fontdimen3\font minus \fontdimen4\font\relax}
\providecommand{\BIBforeignlanguage}[2]{{%
\expandafter\ifx\csname l@#1\endcsname\relax
\typeout{** WARNING: IEEEtran.bst: No hyphenation pattern has been}%
\typeout{** loaded for the language `#1'. Using the pattern for}%
\typeout{** the default language instead.}%
\else
\language=\csname l@#1\endcsname
\fi
#2}}
\providecommand{\BIBdecl}{\relax}
\BIBdecl

\bibitem{lin2011perceptual}
W.~Lin and C.-C.~J. Kuo, ``Perceptual visual quality metrics: {A} survey,'' \emph{Journal of Visual Communication and Image Representation}, vol.~22, no.~4, pp. 297--312, 2011.

\bibitem{zhai2020perceptual}
G.~Zhai and X.~Min, ``Perceptual image quality assessment: {A} survey,'' \emph{Science China Information Sciences}, vol.~63, pp. 1--52, 2020.

\bibitem{duanmu2021quantifying}
Z.~Duanmu, W.~Liu, Z.~Wang, and Z.~Wang, ``Quantifying visual image quality: {A} bayesian view,'' \emph{Annual Review of Vision Science}, vol.~7, no.~1, pp. 437--464, 2021.

\bibitem{yang2015perceptual}
H.~Yang, Y.~Fang, and W.~Lin, ``Perceptual quality assessment of screen content images,'' \emph{IEEE Transactions on Image Processing}, vol.~24, no.~11, pp. 4408--4421, 2015.

\bibitem{gu2015analysis}
K.~Gu, G.~Zhai, W.~Lin, and M.~Liu, ``The analysis of image contrast: {F}rom quality assessment to automatic enhancement,'' \emph{IEEE Transactions on Cybernetics}, vol.~46, no.~1, pp. 284--297, 2015.

\bibitem{li2017quality}
L.~Li, Y.~Zhou, K.~Gu, W.~Lin, and S.~Wang, ``Quality assessment of {DIBR}-synthesized images by measuring local geometric distortions and global sharpness,'' \emph{IEEE Transactions on Multimedia}, vol.~20, no.~4, pp. 914--926, 2017.

\bibitem{zhang2021plug}
K.~Zhang, Y.~Li, W.~Zuo, L.~Zhang, L.~Van~Gool, and R.~Timofte, ``Plug-and-play image restoration with deep denoiser prior,'' \emph{IEEE Transactions on Pattern Analysis and Machine Intelligence}, vol.~44, no.~10, pp. 6360--6376, 2021.

\bibitem{wang2004image}
Z.~Wang, A.~C. Bovik, H.~R. Sheikh, and E.~P. Simoncelli, ``Image quality assessment: {F}rom error visibility to structural similarity,'' \emph{IEEE Transactions on Image Processing}, vol.~13, no.~4, pp. 600--612, 2004.

\bibitem{zhang2018unreasonable}
R.~Zhang, P.~Isola, A.~A. Efros, E.~Shechtman, and O.~Wang, ``The unreasonable effectiveness of deep features as a perceptual metric,'' in \emph{IEEE Conference on Computer Vision and Pattern Recognition}, 2018, pp. 586--595.

\bibitem{ding2020image}
K.~Ding, K.~Ma, S.~Wang, and E.~P. Simoncelli, ``Image quality assessment: {U}nifying structure and texture similarity,'' \emph{IEEE Transactions on Pattern Analysis and Machine Intelligence}, vol.~44, no.~5, pp. 2567--2581, 2020.

\bibitem{ADISTS}
K.~Ding, Y.~Liu, X.~Zou, S.~Wang, and K.~Ma, ``Locally adaptive structure and texture similarity for image quality assessment,'' \emph{ACM International Conference on Multimedia}, pp. 2483--2491, 2021.

\bibitem{heusel2017gans}
M.~Heusel, H.~Ramsauer, T.~Unterthiner, B.~Nessler, and S.~Hochreiter, ``{GAN}s trained by a two time-scale update rule converge to a local nash equilibrium,'' in \emph{Neural Information Processing Systems}, 2017, pp. 1--9.

\bibitem{kligvasser2021deep}
I.~Kligvasser, T.~Shaham, Y.~Bahat, and T.~Michaeli, ``Deep self-dissimilarities as powerful visual fingerprints,'' in \emph{Neural Information Processing Systems}, 2021, pp. 3939--3951.

\bibitem{liao2022deepwsd}
X.~Liao, B.~Chen, H.~Zhu, S.~Wang, M.~Zhou, and S.~Kwong, ``Deep{WSD}: {P}rojecting degradations in perceptual space to wasserstein distance in deep feature space,'' in \emph{ACM International Conference on Multimedia}, 2022.

\bibitem{Manoj2022Do}
\BIBentryALTinterwordspacing
M.~Kumar, N.~Houlsby, N.~Kalchbrenner, and E.~D. Cubuk, ``Do better imagenet classifiers assess perceptual similarity better?'' in \emph{Transactions on Machine Learning Research}, 2022. [Online]. Available: \url{https://openreview.net/forum?id=qrGKGZZvH0}
\BIBentrySTDinterwordspacing

\bibitem{geirhos2018imagenet}
R.~Geirhos, P.~Rubisch, C.~Michaelis, M.~Bethge, F.~A. Wichmann, and W.~Brendel, ``Image{N}et-trained {CNN}s are biased towards texture; increasing shape bias improves accuracy and robustness,'' in \emph{International Conference on Learning Representations}, 2018, pp. 1--22.

\bibitem{hermann2020origins}
K.~Hermann, T.~Chen, and S.~Kornblith, ``The origins and prevalence of texture bias in convolutional neural networks,'' \emph{Neural Information Processing Systems}, vol.~33, pp. 19\,000--19\,015, 2020.

\bibitem{deng2009imagenet}
J.~Deng, W.~Dong, R.~Socher, L.-J. Li, K.~Li, and F.-F. Li, ``Image{N}et: {A} large-scale hierarchical image database,'' in \emph{IEEE Conference on Computer Vision and Pattern Recognition}, 2009, pp. 248--255.

\bibitem{gatys2015texture}
L.~Gatys, A.~S. Ecker, and M.~Bethge, ``Texture synthesis using convolutional neural networks,'' in \emph{Neural Information Processing Systems}, 2015, pp. 1--8.

\bibitem{heitz2021sliced}
E.~Heitz, K.~Vanhoey, T.~Chambon, and L.~Belcour, ``A sliced wasserstein loss for neural texture synthesis,'' in \emph{IEEE Conference on Computer Vision and Pattern Recognition}, 2021, pp. 9412--9420.

\bibitem{navarrete2018multi}
P.~Navarrete~Michelini, D.~Zhu, and H.~Liu, ``Multi--scale recursive and perception--distortion controllable image super--resolution,'' in \emph{European Conference on Computer Vision Workshops}, 2018, pp. 1--14.

\bibitem{julesz1962visual}
B.~Julesz, ``Visual pattern discrimination,'' \emph{IRE Transactions on Information Theory}, vol.~8, no.~2, pp. 84--92, 1962.

\bibitem{portilla2000parametric}
J.~Portilla and E.~P. Simoncelli, ``A parametric texture model based on joint statistics of complex wavelet coefficients,'' \emph{International journal of computer vision}, vol.~40, pp. 49--70, 2000.

\bibitem{szekely2007measuring}
G.~J. Sz{\'e}kely, M.~L. Rizzo, and N.~K. Bakirov, ``Measuring and testing dependence by correlation of distances,'' \emph{The Annals of Statistics}, vol.~35, no.~6, pp. 2769--2794, 2007.

\bibitem{szekely2023energy}
G.~J. Sz{\'e}kely and M.~L. Rizzo, \emph{The energy of data and distance correlation}.\hskip 1em plus 0.5em minus 0.4em\relax CRC Press, 2023.

\bibitem{dokmanic2015euclidean}
I.~Dokmanic, R.~Parhizkar, J.~Ranieri, and M.~Vetterli, ``Euclidean distance matrices: {E}ssential theory, algorithms, and applications,'' \emph{IEEE Signal Processing Magazine}, vol.~32, no.~6, pp. 12--30, 2015.

\bibitem{szekely2009brownian}
G.~J. Sz{\'e}kely and M.~L. Rizzo, ``Brownian distance covariance,'' \emph{The Annals of Applied Statistics}, vol.~3, no.~4, pp. 1236--1265, 2009.

\bibitem{mantiuk2011hdr}
R.~Mantiuk, K.~J. Kim, A.~G. Rempel, and W.~Heidrich, ``{HDR-VDP-2: A} calibrated visual metric for visibility and quality predictions in all luminance conditions,'' \emph{ACM Transactions on Graphics}, vol.~30, no.~4, pp. 1--14, 2011.

\bibitem{xue2013gradient}
W.~Xue, L.~Zhang, X.~Mou, and A.~C. Bovik, ``Gradient magnitude similarity deviation: {A} highly efficient perceptual image quality index,'' \emph{IEEE Transactions on Image Processing}, vol.~23, no.~2, pp. 684--695, 2013.

\bibitem{fang2017objective}
Y.~Fang, J.~Yan, J.~Liu, S.~Wang, Q.~Li, and Z.~Guo, ``Objective quality assessment of screen content images by uncertainty weighting,'' \emph{IEEE Transactions on Image Processing}, vol.~26, no.~4, pp. 2016--2027, 2017.

\bibitem{min2017unified}
X.~Min, K.~Ma, K.~Gu, G.~Zhai, Z.~Wang, and W.~Lin, ``Unified blind quality assessment of compressed natural, graphic, and screen content images,'' \emph{IEEE Transactions on Image Processing}, vol.~26, no.~11, pp. 5462--5474, 2017.

\bibitem{wang2016just}
S.~Wang, L.~Ma, Y.~Fang, W.~Lin, S.~Ma, and W.~Gao, ``Just noticeable difference estimation for screen content images,'' \emph{IEEE Transactions on Image Processing}, vol.~25, no.~8, pp. 3838--3851, 2016.

\bibitem{mannos1974effects}
J.~Mannos and D.~Sakrison, ``The effects of a visual fidelity criterion of the encoding of images,'' \emph{IEEE Transactions on Information Theory}, vol.~20, no.~4, pp. 525--536, 1974.

\bibitem{lin2003discriminative}
W.~Lin, D.~Li, and P.~Xue, ``Discriminative analysis of pixel difference towards picture quality prediction,'' in \emph{International Conference on Image Processing}, 2003, pp. 193--196.

\bibitem{wang2009mean}
Z.~Wang and A.~C. Bovik, ``Mean squared error: {L}ove it or leave it? {A} new look at signal fidelity measures,'' \emph{IEEE Signal Processing Magazine}, vol.~26, no.~1, pp. 98--117, 2009.

\bibitem{watson1993dctune}
A.~B. Watson, ``{DCT}une: {A} technique for visual optimization of dct quantization matrices for individual images,'' in \emph{International Symposium Digest of Technical Papers}, 1993, pp. 946--946.

\bibitem{malo1997subjective}
J.~Malo, A.~Pons, and J.~M. Artigas, ``Subjective image fidelity metric based on bit allocation of the human visual system in the {DCT} domain,'' \emph{Image and Vision Computing}, vol.~15, no.~7, pp. 535--548, 1997.

\bibitem{teo1994perceptual}
P.~C. Teo and D.~J. Heeger, ``Perceptual image distortion,'' in \emph{International Conference on Image Processing}, 1994, pp. 982--986.

\bibitem{watson1997visibility}
A.~B. Watson, G.~Y. Yang, J.~A. Solomon, and J.~Villasenor, ``Visibility of wavelet quantization noise,'' \emph{IEEE Transactions on image processing}, vol.~6, no.~8, pp. 1164--1175, 1997.

\bibitem{larson2010most}
E.~C. Larson and D.~M. Chandler, ``Most apparent distortion: {F}ull-reference image quality assessment and the role of strategy,'' \emph{Journal of electronic imaging}, vol.~19, no.~1, pp. 1--21, 2010.

\bibitem{laparra2016perceptual}
V.~Laparra, J.~Ball{\'e}, A.~Berardino, and E.~P. Simoncelli, ``Perceptual image quality assessment using a normalized laplacian pyramid,'' \emph{Electronic Imaging}, vol. 2016, no.~16, pp. 1--6, 2016.

\bibitem{wang2003multiscale}
Z.~Wang, E.~P. Simoncelli, and A.~C. Bovik, ``Multiscale structural similarity for image quality assessment,'' in \emph{Asilomar Conference on Signals, Systems \& Computers}, 2003, pp. 1398--1402.

\bibitem{wang2010information}
Z.~Wang and Q.~Li, ``Information content weighting for perceptual image quality assessment,'' \emph{IEEE Transactions on Image Processing}, vol.~20, no.~5, pp. 1185--1198, 2010.

\bibitem{wang2005translation}
Z.~Wang and E.~P. Simoncelli, ``Translation insensitive image similarity in complex wavelet domain,'' in \emph{IEEE International Conference on Acoustics, Speech, and Signal Processing}, 2005, pp. 573--576.

\bibitem{zhang2011fsim}
L.~Zhang, L.~Zhang, X.~Mou, and D.~Zhang, ``{FSIM}: {A} feature similarity index for image quality assessment,'' \emph{IEEE Transactions on Image Processing}, vol.~20, no.~8, pp. 2378--2386, 2011.

\bibitem{zhang2014vsi}
L.~Zhang, Y.~Shen, and H.~Li, ``{VSI}: {A} visual saliency-induced index for perceptual image quality assessment,'' \emph{IEEE Transactions on Image processing}, vol.~23, no.~10, pp. 4270--4281, 2014.

\bibitem{sheikh2005information}
H.~R. Sheikh, A.~C. Bovik, and G.~De~Veciana, ``An information fidelity criterion for image quality assessment using natural scene statistics,'' \emph{IEEE Transactions on Image Processing}, vol.~14, no.~12, pp. 2117--2128, 2005.

\bibitem{sheikh2006image}
H.~R. Sheikh and A.~C. Bovik, ``Image information and visual quality,'' \emph{IEEE Transactions on Image Processing}, vol.~15, no.~2, pp. 430--444, 2006.

\bibitem{chang2013sparse}
H.-W. Chang, H.~Yang, Y.~Gan, and M.-H. Wang, ``Sparse feature fidelity for perceptual image quality assessment,'' \emph{IEEE Transactions on Image Processing}, vol.~22, no.~10, pp. 4007--4018, 2013.

\bibitem{wang2006quality}
Z.~Wang, G.~Wu, H.~R. Sheikh, E.~P. Simoncelli, E.-H. Yang, and A.~C. Bovik, ``Quality-aware images,'' \emph{IEEE Transactions on Image Processing}, vol.~15, no.~6, pp. 1680--1689, 2006.

\bibitem{soundararajan2011rred}
R.~Soundararajan and A.~C. Bovik, ``{RRED} indices: {R}educed reference entropic differencing for image quality assessment,'' \emph{IEEE Transactions on Image Processing}, vol.~21, no.~2, pp. 517--526, 2011.

\bibitem{lin2016cross}
L.~Lin, G.~Wang, W.~Zuo, X.~Feng, and L.~Zhang, ``Cross-domain visual matching via generalized similarity measure and feature learning,'' \emph{IEEE Transactions on Pattern Analysis and Machine Intelligence}, vol.~39, no.~6, pp. 1089--1102, 2016.

\bibitem{liu2012image}
T.-J. Liu, W.~Lin, and C.-C.~J. Kuo, ``Image quality assessment using multi-method fusion,'' \emph{IEEE Transactions on Image Processing}, vol.~22, no.~5, pp. 1793--1807, 2012.

\bibitem{simonyan2014very}
K.~Simonyan and A.~Zisserman, ``Very deep convolutional networks for large-scale image recognition,'' in \emph{International Conference on Learning Representations}, 2015, pp. 1--6.

\bibitem{gao2017deepsim}
F.~Gao, Y.~Wang, P.~Li, M.~Tan, J.~Yu, and Y.~Zhu, ``Deep{S}im: {D}eep similarity for image quality assessment,'' \emph{Neurocomputing}, vol. 257, pp. 104--114, 2017.

\bibitem{ghildyal2022shift}
A.~Ghildyal and F.~Liu, ``Shift-tolerant perceptual similarity metric,'' in \emph{European Conference on Computer Vision}, 2022, pp. 91--107.

\bibitem{kim2017deep}
J.~Kim and S.~Lee, ``Deep learning of human visual sensitivity in image quality assessment framework,'' in \emph{IEEE Conference on Computer Vision and Pattern Recognition}, 2017, pp. 1676--1684.

\bibitem{bosse2017deep}
S.~Bosse, D.~Maniry, K.-R. M{\"u}ller, T.~Wiegand, and W.~Samek, ``Deep neural networks for no-reference and full-reference image quality assessment,'' \emph{IEEE Transactions on Image Processing}, vol.~27, no.~1, pp. 206--219, 2017.

\bibitem{prashnani2018pieapp}
E.~Prashnani, H.~Cai, Y.~Mostofi, and P.~Sen, ``Pie{APP}: {P}erceptual image-error assessment through pairwise preference,'' in \emph{IEEE Conference on Computer Vision and Pattern Recognition}, 2018, pp. 1808--1817.

\bibitem{bhardwaj2020unsupervised}
S.~Bhardwaj, I.~Fischer, J.~Ball{\'e}, and T.~Chinen, ``An unsupervised information-theoretic perceptual quality metric,'' in \emph{Neural Information Processing Systems}, 2020, pp. 13--24.

\bibitem{cao2022incorporating}
Y.~Cao, Z.~Wan, D.~Ren, Z.~Yan, and W.~Zuo, ``Incorporating semi-supervised and positive-unlabeled learning for boosting full reference image quality assessment,'' in \emph{IEEE Conference on Computer Vision and Pattern Recognition}, 2022, pp. 5851--5861.

\bibitem{abramowitz1988handbook}
M.~Abramowitz, I.~A. Stegun, and R.~H. Romer, \emph{Handbook of mathematical functions with formulas, graphs, and mathematical tables}.\hskip 1em plus 0.5em minus 0.4em\relax American Association of Physics Teachers, 1988.

\bibitem{wang2008maximum}
Z.~Wang and E.~P. Simoncelli, ``Maximum differentiation ({MAD}) competition: {A} methodology for comparing computational models of perceptual quantities,'' \emph{Journal of Vision}, vol.~8, no.~12, pp. 1--13, 2008.

\bibitem{berardino2017eigen}
A.~Berardino, V.~Laparra, J.~Ball{\'e}, and E.~Simoncelli, ``Eigen-distortions of hierarchical representations,'' in \emph{Neural Information Processing Systems}, 2017, pp. 3530--3539.

\bibitem{simoncelli1995steerable}
E.~P. Simoncelli and W.~T. Freeman, ``The steerable pyramid: {A} flexible architecture for multi-scale derivative computation,'' in \emph{IEEE International Conference on Image Processing}, 1995, pp. 444--447.

\bibitem{zhu2022learing}
H.~Zhu, B.~Chen, l.~Zhu, and S.~Wang, ``Learning spatiotemporal interactions for user-generated video quality assessment,'' \emph{IEEE Transactions on Circuits and Systems for Video Technology}, vol.~33, no.~3, pp. 1031--1042, 2023.

\bibitem{sheikh2003image}
\BIBentryALTinterwordspacing
H.~R. Sheikh, Z.~Wang, L.~Cormack, and A.~C. Bovik, ``Image and video quality assessment research at {LIVE},'' 2006. [Online]. Available: \url{https://live.ece.utexas.edu/research/Quality/subjective.htm}
\BIBentrySTDinterwordspacing

\bibitem{ponomarenko2015image}
N.~Ponomarenko, L.~Jin, O.~Ieremeiev, V.~Lukin, K.~Egiazarian, J.~Astola, B.~Vozel, K.~Chehdi, M.~Carli, F.~Battisti \emph{et~al.}, ``Image database {TID2013}: {P}eculiarities, results and perspectives,'' \emph{Signal Processing: Image Communication}, vol.~30, pp. 57--77, 2015.

\bibitem{2019KADID}
H.~Lin, V.~Hosu, and D.~Saupe, ``{KADID}-10k: {A} large-scale artificially distorted {IQA} database,'' in \emph{International Conference on Quality of Multimedia Experience}, 2019, pp. 1--3.

\bibitem{jinjin2020pipal}
G.~Jinjin, C.~Haoming, C.~Haoyu, Y.~Xiaoxing, J.~S. Ren, and D.~Chao, ``{PIPAL}: {A} large-scale image quality assessment dataset for perceptual image restoration,'' in \emph{European Conference on Computer Vision}, 2020, pp. 633--651.

\bibitem{video2000final}
\BIBentryALTinterwordspacing
VQEG, ``{Final} report from the video quality experts group on the validation of objective models of video quality assessment,'' 2000. [Online]. Available: \url{http://www.vqeg.org}
\BIBentrySTDinterwordspacing

\bibitem{golestaneh2015effect}
S.~A. Golestaneh, M.~M. Subedar, and L.~J. Karam, ``The effect of texture granularity on texture synthesis quality,'' \emph{Applications of Digital Image Processing XXXVIII}, vol. 9599, pp. 356--361, 2015.

\bibitem{zujovic2013structural}
J.~Zujovic, T.~N. Pappas, and D.~L. Neuhoff, ``Structural texture similarity metrics for image analysis and retrieval,'' \emph{IEEE Transactions on Image Processing}, vol.~22, no.~7, pp. 2545--2558, 2013.

\bibitem{alfarraj2016content}
M.~Alfarraj, Y.~Alaudah, and G.~AlRegib, ``Content-adaptive non-parametric texture similarity measure,'' in \emph{IEEE International Workshop on Multimedia Signal Processing}, 2016, pp. 1--6.

\bibitem{golestaneh2018synthesized}
A.~Golestaneh and L.~J. Karam, ``Synthesized texture quality assessment via multi-scale spatial and statistical texture attributes of image and gradient magnitude coefficients,'' in \emph{IEEE Conference on Computer Vision and Pattern Recognition Workshops}, 2018, pp. 738--744.

\bibitem{snelgrove2017high}
X.~Snelgrove, ``High-resolution multi-scale neural texture synthesis,'' in \emph{SIGGRAPH Asia Technical Briefs}, 2017, pp. 1--4.

\bibitem{mcinnes2018umap}
L.~McInnes, J.~Healy, N.~Saul, and L.~Grossberger, ``{UMAP}: {U}niform manifold approximation and projection,'' \emph{The Journal of Open Source Software}, vol.~3, no.~29, pp. 861--862, 2018.

\bibitem{he2016deep}
K.~He, X.~Zhang, S.~Ren, and J.~Sun, ``Deep residual learning for image recognition,'' in \emph{IEEE Conference on Computer Vision and Pattern Recognition}, 2016, pp. 770--778.

\bibitem{huang2017densely}
G.~Huang, Z.~Liu, L.~Van Der~Maaten, and K.~Q. Weinberger, ``Densely connected convolutional networks,'' in \emph{IEEE Conference on Computer Vision and Pattern Recognition}, 2017, pp. 4700--4708.

\bibitem{mechrez2018contextual}
R.~Mechrez, I.~Talmi, and L.~Zelnik-Manor, ``The contextual loss for image transformation with non-aligned data,'' in \emph{European Conference on Computer Vision}, 2018, pp. 768--783.

\bibitem{guo2010completed}
Z.~Guo, L.~Zhang, and D.~Zhang, ``A completed modeling of local binary pattern operator for texture classification,'' \emph{IEEE Transactions on Image Processing}, vol.~19, no.~6, pp. 1657--1663, 2010.

\bibitem{dong2020perceptual}
X.~Dong, J.~Dong, and M.~J. Chantler, ``Perceptual texture similarity estimation: An evaluation of computational features,'' \emph{IEEE Transactions on Pattern Analysis and Machine Intelligence}, vol.~43, no.~7, pp. 2429--2448, 2020.

\bibitem{wei2000fast}
L.-Y. Wei and M.~Levoy, ``Fast texture synthesis using tree-structured vector quantization,'' in \emph{Annual Conference on Computer Graphics and Interactive Techniques}, 2000, pp. 479--488.

\bibitem{efros1999texture}
A.~A. Efros and T.~K. Leung, ``Texture synthesis by non-parametric sampling,'' in \emph{IEEE International Conference on Computer Vision}, 1999, pp. 1033--1038.

\bibitem{kwatra2005texture}
V.~Kwatra, I.~Essa, A.~Bobick, and N.~Kwatra, ``Texture optimization for example-based synthesis,'' in \emph{ACM SIGGRAPH}, 2005, pp. 795--802.

\bibitem{Tsukida2011}
K.~Tsukida and M.~R. Gupta, ``How to analyze paired comparison data,'' {T}echnical Report UWEETR-2011-0004, University of Washington, May 2011.

\bibitem{byrd1995limited}
R.~H. Byrd, P.~Lu, J.~Nocedal, and C.~Zhu, ``A limited memory algorithm for bound constrained optimization,'' \emph{SIAM Journal on Scientific Computing}, vol.~16, no.~5, pp. 1190--1208, 1995.

\bibitem{johnson2016perceptual}
J.~Johnson, A.~Alahi, and L.~Fei-Fei, ``Perceptual losses for real-time style transfer and super-resolution,'' in \emph{European Conference on Computer Vision}, 2016, pp. 694--711.

\bibitem{huang2017arbitrary}
X.~Huang and S.~Belongie, ``Arbitrary style transfer in real-time with adaptive instance normalization,'' in \emph{IEEE International Conference on Computer Vision}, 2017, pp. 1501--1510.

\bibitem{li2022sliced}
J.~Li, D.~Xu, and S.~Yao, ``Sliced wasserstein distance for neural style transfer,'' \emph{Computers \& Graphics}, vol. 102, pp. 89--98, 2022.

\bibitem{jing2019neural}
Y.~Jing, Y.~Yang, Z.~Feng, J.~Ye, Y.~Yu, and M.~Song, ``Neural style transfer: {A} review,'' \emph{IEEE Transactions on Vsualization and Computer Graphics}, vol.~26, no.~11, pp. 3365--3385, 2019.

\bibitem{li2017universal}
Y.~Li, C.~Fang, J.~Yang, Z.~Wang, X.~Lu, and M.-H. Yang, ``Universal style transfer via feature transforms,'' \emph{Neural Information Processing Systems}, vol.~30, 2017.

\bibitem{risser2017stable}
E.~Risser, P.~Wilmot, and C.~Barnes, ``Stable and controllable neural texture synthesis and style transfer using histogram losses,'' \emph{arXiv preprint arXiv:1701.08893}, 2017.

\bibitem{zhang2022exact}
Y.~Zhang, M.~Li, R.~Li, K.~Jia, and L.~Zhang, ``Exact feature distribution matching for arbitrary style transfer and domain generalization,'' in \emph{IEEE Conference on Computer Vision and Pattern Recognition}, 2022, pp. 8035--8045.

\bibitem{kotovenko2019content}
D.~Kotovenko, A.~Sanakoyeu, S.~Lang, and B.~Ommer, ``Content and style disentanglement for artistic style transfer,'' in \emph{IEEE International Conference on Computer Vision}, 2019, pp. 4422--4431.

\bibitem{azadi2018multi}
S.~Azadi, M.~Fisher, V.~G. Kim, Z.~Wang, E.~Shechtman, and T.~Darrell, ``Multi-content {GAN} for few-shot font style transfer,'' in \emph{IEEE Conference on Computer Vision and Pattern Recognition}, 2018, pp. 7564--7573.

\bibitem{lin2014microsoft}
T.-Y. Lin, M.~Maire, S.~Belongie, J.~Hays, P.~Perona, D.~Ramanan, P.~Doll{\'a}r, and C.~L. Zitnick, ``Microsoft {COCO}: {C}ommon objects in context,'' in \emph{European Conference on Computer Vision}, 2014, pp. 740--755.

\bibitem{wikiart2016}
\BIBentryALTinterwordspacing
K.~Nichol, ``Painter by numbers, {W}ikiart,'' 2016. [Online]. Available: \url{https://www.kaggle.com/c/painter-by-numbers}
\BIBentrySTDinterwordspacing

\bibitem{Kingma2014adam}
D.~P. Kingma and J.~Ba, ``Adam: {A} method for stochastic optimization,'' \emph{CoRR}, vol. abs/1412.6980, 2014.

\end{thebibliography}

\end{document}